\documentclass[journal]{IEEEtran} 
\usepackage{subfigure}
\usepackage{amsmath,amssymb}
\usepackage{url,booktabs}
\usepackage{graphicx}
\usepackage{flushend}
\usepackage{bm}
\usepackage[usenames,dvipsnames]{xcolor}
\usepackage{cite}

\RequirePackage{snapshot}

\begin{document}

\title{%
Bias Correction
and
Modified Profile Likelihood
under the Wishart Complex Distribution}

\author{Abra\~ao~D.~C.~Nascimento,
        Alejandro~C.~Frery,~\IEEEmembership{Senior Member}, 
        and
        Renato~J.~Cintra,~\IEEEmembership{Senior Member}
\thanks{This work was supported by CNPq, Fapeal and FACEPE, Brazil.}
\thanks{A.~D.~C.~Nascimento is with the 
Departamento de Estat\'istica, 
Universidade Federal de Pernambuco, 
Cidade Universit\'aria, 50740-540, Recife, PE, Brazil, 
e-mail: abraao.susej@gmail.com}

\thanks{This work was partially funded by Conicet, CNPq, Fapeal and Capes.}
\thanks{
A.~C.~Frery is with the LaCCAN -- Laborat\'orio de Computa\c c\~ao Cient\'ifica e An\'alise Num\'erica, Universidade Federal de Alagoas, BR 104 Norte km 97, 57072-970, Macei\'o, AL, Brazil, e-mail: acfrery@gmail.com}
\thanks{%
R.~J.~Cintra is with the Signal Processing Group, 
Departamento de Estat\'istica, 
Universidade Federal de Pernambuco, 
Cidade Universit\'aria, 
50740-540, Recife, PE, Brazil, 
e-mail: rjdsc@ieee.org}
}

\markboth{ }
{Nascimento \MakeLowercase{\textit{et al.}}: Nascimento}

\maketitle

\begin{abstract}
This paper proposes improved methods for the maximum likelihood (ML) estimation of the equivalent number of looks $L$.
This parameter has a meaningful interpretation in the context of polarimetric synthetic aperture radar (PolSAR) images.
Due to the presence of coherent illumination in their processing, PolSAR systems generate images which present a granular noise called speckle.
As a potential solution for reducing such interference, the parameter $L$ controls the signal-noise ratio.
Thus, the proposal of efficient estimation methodologies for $L$ has been sought.
To that end, we consider firstly that a PolSAR image is well described by the scaled complex Wishart distribution.
In recent years, 
%Anfinsen~\textit{et al.}~[IEEE Transactions on Geoscience and Remote Sensing, vol. 47, no. 11, pp. 3795--3809, 2009] 
Anfinsen~\textit{et al.}
%~\cite{EstimationEquivalentNumberLooksSAR}
derived and analyzed estimation methods based on the ML and on trace statistical moments for obtaining the parameter $L$ of the unscaled version of such probability law.
This paper generalizes that approach.
We present the second-order bias expression proposed by 
%Cox and Snell~[Journal of the Royal Statistical Society. Series B (Methodological), vol. 30, no. 2, pp. 248--275, 1968] 
Cox and Snell
%~\cite{CoxandSnell1968}
for the ML estimator of this parameter.
Moreover, the formula of the profile likelihood modified by 
%Barndorff-Nielsen~[Journal of the Royal Statistical Society. Series B (Methodological), vol. 56, no. 1, pp. 125--140, 1994] 
Barndorff-Nielsen
%~\cite{BarndorffNielsen1994}
in terms of $L$ is discussed.
Such derivations yield two new ML estimators for the parameter $L$, which are compared to the estimators proposed by Anfinsen~\textit{et al.}.
The performance of these estimators is assessed by means of Monte Carlo experiments, adopting three statistical measures as comparison criterion: the mean square error, the bias, and the coefficient of variation.
Equivalently to the simulation study, an application to actual PolSAR data concludes that the proposed estimators outperform all the others in homogeneous scenarios.
\end{abstract}

\section{Introduction}\label{correction:intro}

Multilook polarimetric synthetic aperture radar (PolSAR) is a technology which has proven to be an effective tool for geophysical remote sensing~\cite{LeePottier2009PolarimetricRadarImaging}.
%This systems work on the sending of electromagnetic pulses with several polarization channels toward a target and the capture of the returning echos.
This particular system operates by sending electromagnetic pulses of several polarizations towards 
a remote target; then the returning echoes are captured and registered.
This process is characterized by the use of coherent illumination,
which 
affects
the obtained images with a granular noise called `speckle'.
Therefore, a specialized modeling is required for processing and analyzing such images.

In this context, 
the scaled complex Wishart distribution 
has been suggested as a successful approach 
for modeling the backscattered PolSAR signal.
This distribution is equipped with two parameters: 
a covariance matrix and the number of looks ($L$).  
The former quantity
is directly related to the data structure 
and is defined over the set of positive-definite Hermitian matrices~\cite{FreitasFreryCorreia:Environmetrics:03}.
On the other hand, 
parameter $L$ is defined as the mean number of statistically independent samples represented by each pixel.
Indeed, 
parameter $L$ is related to the signal-to-noise ratio
and
the
speckle noise is less severe for large values of $L$~\cite{FreryCintraNascimento2013,Nascimentoetaieee12013}.

Several studies have addressed the issue of 
estimating~$L$~\cite{EstimationEquivalentNumberLooksSAR,GierullSikaneta2002}.
Despite the various existing approaches,
the maximum likelihood (ML) method 
is often selected as the method of choice.
This is mainly
due to its good asymptotic properties,
such as 
consistency and 
convergence in distribution to the Gaussian law~\cite{caselaberge2002}.

However,
the ML estimator
in finite samples
is usually biased with respect to the true parameter value.
Indeed, the ML estimator bias has order $O(N^{-1})$, where $N$ is the sample size and $O(\cdot)$ is Landau notation to represent order.
In practice, this quantity is not commonly considered because its value is negligible when compared to the standard error, which has order $O(N^{-1/2})$. 
However, such biases can be substantial in situations where small or moderate sample sizes are used.
Thus, 
analytic expressions for 
the ML estimator bias 
have been sought in order to derive 
a more accurate,
corrected estimator for finite sample sizes~\cite{VasconcellosandFreryandSilva2005,Cordeiro2011}.

In~\cite{CoxandSnell1968},
Cox and Snell proposed a general formula
for the second-order bias of the ML estimator.
Several works have considered this formula
as a mean to 
obtain improved estimators
for parameters of several distributions~\cite{Cordeiro2011}.
Vasconcellos~\textit{et al.}~\cite{VasconcellosandFreryandSilva2005} 
applied it to the univariate $\mathcal G^0$ distribution, 
which is often 
used for modeling 
synthetic aperture radar (SAR) images with one polarization channel~\cite{freryetal1997a}.
Pianto and Cribari-Neto~\cite{Pianto20111394} 
proposed
a bias correction approach
for the roughness parameter of the $\mathcal G^0$ distribution. 

More accurate ML estimators can also be derived by means of
the modified profile likelihood~\cite[chapter 9]{bookseverini2000}.
In this approach, 
an improved parameter estimation 
can be achieved by maximizing 
an adjusted profiled likelihood function~\cite{Fraserandreid}.
Barndorff-Nielsen proposed a successful adjustment for 
the profile likelihood in~\cite{BarndorffNielsen1994}.
This methodology could offer parametric inference techniques
suitable for several distributions.
The Birnbaum-Saunders distribution~\cite{Cysneiros20084939} and elliptical structural models~\cite{FerrariandMelo2010} are illustrative examples.
This approach has also been 
employed in SAR data models.
Silva~\textit{et al.}~\cite{SilvaCribariFrery:ImprovedLikelihood:Environmetrics} proposed point estimation and hypothesis test methods for the roughness parameter of the $\mathcal G^0$ distribution based on modified profile likelihood. 

Anfinsen~\textit{et al.}~\cite{EstimationEquivalentNumberLooksSAR} proposed three methods for estimating the number of looks. 
Their methods were favorably compared with classic methods based on fractional moment and the coefficient of variation.
Assuming that PolSAR images are well modeled by the complex Wishart distribution, 
the ML estimator was found to excel when compared to other classic techniques.
Additionally,
such estimators were submitted to bias analysis
by means of jackknife and bootstrap 
techniques~\cite{EstimationEquivalentNumberLooksSAR}.
However, 
these techniques are computationally intensive and
are prone to inaccuracies 
when small sample sizes are considered~\cite{FreryandCribariNetoandSouza2004}.

The aim of this paper is three-fold.
First,
we derive 
a second-order bias expression for the ML estimator of parameter $L$
and
propose corrected estimator for $L$
following
Cox-Snell approach~\cite{CoxandSnell1968}. 
Second,
considering
the scaled complex Wishart law,
we propose
a new estimator
for $L$ 
according
Barndorff-Nielsen correction~\cite{BarndorffNielsen1994}.
Third,
the performance of the proposed
corrected ML estimators 
is quantified and compared to 
the estimators listed in literature.
This assessment is performed according to
Monte Carlo experiments 
at several different simulation scenarios.
Actual PolSAR data are also considered and analyzed.

The remainder of this paper is structured as follows.
In Section~\ref{correction:model},
we review
the scaled complex Wishart distribution,
which is the probability model considered in this work. 
Section~\ref{correction:methodology} 
discusses
Cox-Snell and Barndorff-Nielsen
methodologies for obtaining corrected ML estimators.
In Section~\ref{correction:results},
the new corrected estimators
are introduced and mathematically described.
Subsequently, 
the derived estimators are analyzed and
compared with existing literature techniques 
in Section~\ref{correction:results}.
Finally, 
conclusions are summarized in Section~\ref{correction:conclusion}. 

\section{Scaled Complex Wishart distribution and ML Estimation}\label{correction:model}

\subsection{Scaled Complex Wishart distribution}

The polarimetric coherent information associates to each image pixel 
a 2$\times$2 complex matrix.
The entries of such matrix are 
$S_\text{VV}$, $S_\text{VH}$, $S_\text{HV}$, and $S_\text{HH}$, 
where $S_{ij}$ represents the backscattered signal 
for the $i$th transmission and $j$th reception linear polarization, 
for $i,j=\text{H},\text{V}$.
Under the reciprocity theorem conditions~\cite{UlabyElachi1990}, 
the scattering matrix is often simplified into a three-component vector, 
because $S_\text{HV}=S_\text{VH}$.
Thus, we have 
the following scattering vector
\begin{align*}
\begin{bmatrix}
S_\text{VV}
&
\sqrt{2}\,S_\text{VH}
&
S_\text{HH}
\end{bmatrix}^\top,
\end{align*}
where the superscript ${}^\top$ denotes vector transposition.
This three-component representation is detailed in~\cite{FreemananDurden}.

In general, we may consider a system with $m$ polarizations. 
Thus the associated complex random vector is denoted by:
\begin{align*}
\bm{s}
=
\begin{bmatrix}
S_1 & S_2 & \cdots & S_m
\end{bmatrix}^\top
,
\end{align*}
where $S_j$, $j=1,2,\ldots,m$, 
represents the random variables 
associated with each polarization channel.
This particular 
random vector 
has been successfully 
modeled by 
the multivariate complex Gaussian distribution 
with zero-mean, as discussed in~\cite{Goodmanb}.
Thus, the single-look PolSAR data representation is 
essentially given by $\bm{s}$.

Multilook polarimetric speckled data obtained 
from scattered signal 
is commonly defined by the
estimated sample covariance matrix~\cite{EstimationEquivalentNumberLooksSAR},
which
is given by
$$
\bm{Z}
=
\frac{1}{L}
\sum_{i=1}^L 
\bm{s}_i \bm{s}_i^{\text{H}},
$$
where the superscript ${}^{\text{H}}$ 
represents the complex conjugate transpose of a vector, 
$\bm{s}_i$, $i=1,2,\ldots,L$, are scattering vectors, 
and $L$ is the number of looks.
Matrix~$\bm{Z}$ is
Hermitian positive definite 
and 
follows a scaled complex Wishart distribution~\cite{EstimationEquivalentNumberLooksSAR}. 
Having $\bm{\Sigma}$ and $L$ as parameters, 
the scaled complex Wishart distribution is associated to the following probability density function~\cite{FreitasFreryCorreia:Environmetrics:03}:
\begin{align*}
\label{eq:denswishart}
f_{\bm{Z}}(\bm{Z}';\bm{\Sigma},L) 
=
\frac{L^{mL}|\bm{Z}'|^{L-m}}{|\bm{\Sigma}|^L \Gamma_m(L)} 
\exp\big(
-
L
\operatorname{tr}(\bm{\Sigma}^{-1}\bm{Z}')
\big)
,
\end{align*}
where 
$\Gamma_m(L)=\pi^{m(m-1)/2}\prod_{i=0}^{m-1}\Gamma(L-i)$ 
for $L\geq m$, 
$\Gamma(\cdot)$ is the gamma function, 
$\operatorname{tr}(\cdot)$ represents the trace operator, 
$|\cdot|$ denotes the determinant operator, 
$\bm{\Sigma}$ is the covariance matrix
associated to~$\bm{s}$,
\begin{align*}
\Sigma=&\operatorname{E}\bigl(\bm{s}\bm{s}^{*}\bigr)\\ 
=&{\left[\begin{array}{cccc}
\operatorname{E}(S_1^{}S_1^{*}) & \operatorname{E}(S_1^{}S_2^{*}) & \cdots &\operatorname{E}(S_1^{}S_m^{*}) \\
\operatorname{E}(S_2^{}S_1^{*}) & \operatorname{E}(S_2^{}S_2^{*}) & \cdots &\operatorname{E}(S_2^{}S_m^{*}) \\
\vdots &  \vdots & \ddots &\vdots \\
\operatorname{E}(S_m^{}S_1^{*}) & \operatorname{E}(S_m^{}S_2^{*}) & \cdots & \operatorname{E}(S_m^{}S_m^{*}) \end{array} \right]},
\end{align*}
and superscript ${}^{*}$ means the conjugate of a complex number. 
The first moment
of $\bm{Z}$
satisfies
$\operatorname{E}(\bm{Z})=\bm{\Sigma}$~\cite{EstimationEquivalentNumberLooksSAR}, where $\operatorname{E}(\cdot)$ is the 
statistical expectation operator.
We denote
$\bm{Z}\sim \mathcal W(\bm{\Sigma},L)$ 
to indicate that $\bm{Z}$
follows the scaled complex Wishart distribution.

\subsection{Estimation of Scaled Complex Wishart Distribution Parameters}\label{ref12}

Let $\{\bm{Z}_1,\bm{Z}_2,\ldots,\bm{Z}_N\}$ 
be a random sample from 
$\bm{Z}\sim \mathcal W(\bm{\Sigma},L)$. 
The ML estimators for $\bm{\Sigma}$ and $L$, 
namely $\widehat{\bm{\Sigma}}_{\text{ML}}$ and $\widehat{L}_{\text{ML}}$, 
respectively, 
are quantities that maximize 
the log-likelihood function associated to 
the Wishart distribution,
which is given by~\cite{EstimationEquivalentNumberLooksSAR,FreryCintraNascimento2012}:
\begin{equation}
\begin{split}
{\ell}(\bm{\theta})
=
&
mNL\log L
+
(L-m)\sum_{k=1}^N
\log|\bm{Z}_k|
-
LN\log|\bm{\Sigma}| 
\\
&
-N\frac{m(m-1)}{2}
\log\pi
-N
\sum_{k=0}^{m-1}
\log\Gamma(L-k)
\\
&
-NL
\operatorname{tr}(\bm{\Sigma}^{-1}\bar{\bm{Z}})
,
\label{likelihood}
\end{split}
\end{equation}
where
$\bar{\bm{Z}}=N^{-1}\,\sum_{k=1}^N\,\bm{Z}_k$,
$\bm{\theta}=\begin{bmatrix}\bm{\sigma}^\top & L \end{bmatrix}^\top$,
$\bm{\sigma}=\operatorname{vec}(\bm{\Sigma})$,
and $\operatorname{vec}(\cdot)$ 
is the column stacking vectorization operator.

In a previous work~\cite{FreryCintraNascimento2012}, 
we have showed that
$\widehat{\bm{\Sigma}}_\text{ML}$
is given by the sample mean,
\begin{align*}
\widehat{\bm{\Sigma}}_\text{ML}=\bar{\bm{Z}}
,
\end{align*}
and $\widehat{L}_{\text{ML}}$ 
satisfies
the following non-linear equation:
\begin{align}
\label{eqscore1}
m\log\widehat{L}_\text{ML}
+
\frac{1}{N}
\sum_{k=1}^N 
\log|\bm{Z}_k|
-
\log|\bar{\bm{Z}}|
-
\psi_m^{(0)}(\widehat{L}_{\text{ML}})
=
0
,
\end{align}
where $\psi_m^{(0)}(\cdot)$ is the 
zeroth order term of the $v$th-order
multivariate polygamma function given by
$$
\psi_m^{(v)}(L)=\sum_{i=0}^{m-1} \psi^{(v)}(L-i),
$$
$\psi^{(v)}(\cdot)$ is the ordinary polygamma function expressed by
$$
\psi^{(v)}(L)=\frac{\partial^{v+1} \log\Gamma(L)}{\partial L^{v+1}},
$$
for $v\geq 0$.
Function $\psi^{(0)}(\cdot)$ is known as the digamma function.

In above expression~\eqref{eqscore1} 
a closed-form analytical solution 
for $\widehat{L}_\text{ML}$~\cite{EstimationEquivalentNumberLooksSAR,FreryCintraNascimento2012}
could not be identified.
Nevertheless,
$\widehat{L}_\text{ML}$
can be obtained via 
the Newton-Raphson numerical
optimization method~\cite{gentle2002elements}.
A detailed account on the performance of
$\widehat{L}_{\text{ML}}$ 
is provided in~\cite{EstimationEquivalentNumberLooksSAR}.
However, 
estimator $\widehat{L}_{\text{ML}}$
is not adequate for applications based on small sample sizes,
such as filtering methods for image restoration~\cite{RestrepoandBovik1988,TorresPolarimetricFilterPatternRecognition}.
%\rjdsc{Nao e' adequado porque?}
This is due to the fact that
the ML estimator is only guaranteed to be asymptotically unbiased;
therefore, 
techniques which depend on ML estimates 
under small samples may
suffer from pronounced bias.

A convenient
estimator based on the method of moments
for the equivalent number of looks (ENL)
was proposed in~\cite{OliverandQuegan1998}
and 
suggested
for
both single-polarization and 
PolSAR data~\cite{EstimationEquivalentNumberLooksSAR}.
However, 
for that particular method,
in the PolSAR case,
the cross-terms of the estimated sample covariance matrix
are not considered~\cite{EstimationEquivalentNumberLooksSAR},
which means that potentially useful information
is simply being discarded.

To address this issue,
Anfinsen~\textit{et al.}~\cite{EstimationEquivalentNumberLooksSAR} 
proposed two estimators 
based on trace moment,
whose expressions are given by:
\begin{align*}
\widehat{L}_\text{MM1}
=
\frac{\operatorname{tr}(\bar{\bm{Z}}\;\bar{\bm{Z}})}
{N^{-1}\sum_{i=1}^N \operatorname{tr}(\bm{Z}_i)^2-\operatorname{tr}(\bar{\bm{Z}})^2}
\end{align*}
and
$$
\widehat{L}_\text{MM2}=\frac{\operatorname{tr}(\bar{\bm{Z}})^2}{N^{-1}\sum_{i=1}^N \operatorname{tr}(\bm{Z}_i\bm{Z}_i)-\operatorname{tr}(\bar{\bm{Z}}\;\bar{\bm{Z}})}.
$$
These estimators are based on the derivations 
proposed by Maiwald and Kraus~\cite{MaiwaldandKraus2000}.

\section{Corrected ML estimation methods}\label{correction:methodology}

Several works have addressed 
the
ML estimator properties in finite sample 
for SAR data modeling~\cite{VasconcellosandFreryandSilva2005,Pianto20111394,SilvaCribariFrery:ImprovedLikelihood:Environmetrics}.
ML estimation for other statistical models 
has also been proposed~\cite{Cordeiro2011,Cysneiros20084939,FerrariandMelo2010}.
In this section,
we describe
two
methodologies for obtaining improved ML estimators.
The first one
is based on the ML estimator second-order bias 
according to the Cox and Snell general formula~\cite{CoxandSnell1968}.
The second method
is based on
the modified profile likelihood 
and
was
proposed by Barndorff-Nielsen~\cite{BarndorffNielsen1994}.

\subsection{Mathematical Framework}

In this subsection,
we present the mathematical framework
required for both
methods above referred to.

Let 
$\{\bm{Z}_1,\bm{Z}_2,\ldots,\bm{Z}_N\}$ 
be a random sample from a random positive-definite Hermitian matrix $\bm{Z}$ 
equipped with  a given density 
$f_{\bm{Z}}(\bm{Z}'; \bm{\theta})$, 
where $\bm{Z}'$ 
represents possible outcomes and
$\bm{\theta} = \begin{bmatrix} \theta_1 & \theta_2 & \cdots & \theta_p\end{bmatrix}^\top$ 
is the parameter vector with length~$p$.

The associated log-likelihood function based on above observations
is given by
\begin{align*}
\ell(\bm{\theta})
=
\sum_{k=1}^N
\log f_{\bm{Z}}(\bm{Z}_k; \bm{\theta})
.
\end{align*}

The derivatives of the log-likelihood play an important role in ML estimation.
For the sake of compactness,
we adopt the following notation:
\begin{align*}
U_{\theta_i}
&=
\frac{\partial}{\partial \theta_i} \ell(\bm{\theta})
,
\\
U_{\theta_i\theta_j}
&=
\frac{\partial^2 }{\partial \theta_i \partial \theta_j}
\ell(\bm{\theta})
,
\\
U_{\theta_i\theta_j\theta_k}
&=
\frac{\partial^3 }{\partial \theta_i \partial \theta_j \partial \theta_k}
\ell(\bm{\theta})
,
\end{align*}
for $i,j,k=1,2,\dots,p$.
Accordingly,
we denote the cumulants for the log-likelihood derivatives
as~\cite{Cordeiro2011}:
\begin{align*}
\kappa_{\theta_i\theta_j}
&=
\operatorname{E}(U_{\theta_i\theta_j})
,
\qquad
\kappa_{\theta_i,\theta_j}
=
\operatorname{E}(U_{\theta_i}U_{\theta_j})
,
\\
\kappa_{\theta_i\theta_j\theta_k}
&=\operatorname{E}(U_{\theta_i\theta_j\theta_k})
,
\qquad
\kappa_{\theta_i,\theta_j\theta_k}
=
\operatorname{E}(U_{\theta_i}U_{\theta_j\theta_k})
.
\end{align*}
As a direct consequence,
the elements of the Fisher information matrix
associated to~$\bm{\theta}$ are
$\kappa_{\theta_i,\theta_j}$.
Additionally, 
we denote the elements of the inverse of 
the Fisher information matrix
as
$\kappa^{\theta_i,\theta_j}$.
%Notice that the element $\kappa_{\theta_i,\theta_j}=-\kappa_{\theta_i\theta_j}$ (known as one of Bartlett identities~\cite{BNandCOX1989}) is an entry of the Fisher information matrix $\mathcal K(\bm{\theta})$  associated to the parameter vector $\bm{\theta}$.
%Additionally, $\kappa^{\theta_i,\theta_j}=-\kappa^{\theta_i\theta_j}$ is an element of $\mathcal K(\bm{\theta})^{-1}$.
%\uline{This last term is strongly utilized in the expression for the bias and follows the notation of Lawley}~\cite{law1956}.

Moreover, 
the first derivative of the cumulants 
with respect to the parameters is denoted by:
\begin{align*}
\kappa_{\theta_i\theta_j}^{(\theta_k)}
=
\frac{\partial}{ \partial \theta_k} 
\kappa_{\theta_i\theta_j}
,
\end{align*}
for $i,j,k=1,2,\dots,p$.

\subsection{Cox-Snell Corrected ML Estimator}
\label{correction:methodology:1}

Let $\hat\theta_i$,
$i=1,2,\ldots,p$,
be the ML estimator for $\theta_i$.
Considering the mathematical expressions described in the previous
subsection,
the
Cox-Snell expression
for the bias of $\hat\theta_i$
is described here.
Such bias correction was given an alternative form
in terms of the cumulants of 
the log-likelihood derivatives~\cite{Cordeiro2011}.
It was established that the bias of~$\hat\theta_i$
possesses the following formulation~\cite{Cordeiro2011}:
\begin{equation}
\label{Bartlettcorrection}
{B}(\widehat\theta_i)
=\operatorname{E}(\widehat\theta_i)-\theta_i=
\sum_{r,s,t} 
\kappa^{\theta_i,\theta_r} 
\kappa^{\theta_s,\theta_t}
\Bigl(
\kappa_{\theta_r\theta_s}^{(\theta_t)}
-
\frac{1}{2} 
\kappa_{\theta_r\theta_s\theta_t}
\Bigr)
,
\end{equation}   
Therefore, 
the corrected ML estimator $\widetilde\theta_i$ 
is defined by 
\begin{equation}
\label{correctedestimator0}
\widetilde\theta_i
=
\widehat\theta_i
-
\widehat{B}(\widehat\theta_i),
\end{equation}
where $\widehat{B}(\widehat\theta_i)$ represents the bias ${B}(\widehat\theta_i)$ evaluated 
not at
$\theta_i$
but
at $\widehat\theta_i$.
Additionally,
it was previously shown that
(i)~$\operatorname{E}[\widehat{B}(\widehat\theta_i)]=O\left(N^{-2}\right)$,
(ii)~$\operatorname{E}(\widehat\theta_i)
=
\theta_i
+
O\left(N^{-1}\right)$,
and
(iii)~$\operatorname{E}(\widetilde\theta_i)
=
\theta_i
+
O\left(N^{-2}\right)$~\cite{CoxandSnell1968}.
Therefore
the bias of $\widetilde\theta_i$ has order of $N^{-2}$.
In contrast, 
we have that 
$B(\widehat\theta_i)=O\left(N^{-1}\right)$~\cite{VasconcellosandFreryandSilva2005}.
Thus, 
$\widetilde\theta_i$ is expected 
to possess better asymptotic properties when compared to $\widehat\theta_i$.

\subsection{Barndorff-Nielsen ML Estimator}
\label{correction:methodology:2}

In~\cite{BarndorffNielsen1994}
Barndorff-Nielsen
introduced an
improved ML estimation
based on
the modified profile likelihood.
This method was previously employed
in~\cite{SilvaCribariFrery:ImprovedLikelihood:Environmetrics,Cysneiros20084939,FerrariandMelo2010}. 
Below we furnish a brief outline of the technique.

Let the parameter vector $\bm{\theta}$
be split into two vectors such that
$\bm{\theta}=\begin{bmatrix} \bm{\psi}^\top & \bm{\lambda}^\top\end{bmatrix}^\top$,
where
$\bm{\psi}$ and $\bm{\lambda}$ have
$q$ and $p-q$ elements,
respectively.
Vector $\bm{\psi}$ is the interest parameter vector,
whereas vector $\bm{\lambda}$ is the nuisance parameter vector.
In~\cite{coxandreid1992,Stern1997}, 
the statistical inference of $\bm{\psi}$,
for a known~$\bm{\lambda}$, was addressed.
However, 
such calculation requires the marginal or 
conditional likelihood functions, 
which are often mathematically intractable~\cite{SilvaCribariFrery:ImprovedLikelihood:Environmetrics}.

A common solution to this conundrum is 
to approximate these functions by the profile log-likelihood~\cite{Fraserandreid},
which is denoted by 
\begin{align*}
\widetilde{\ell}(\bm{\psi})
=
\ell
\left(
\begin{bmatrix}
\bm{\psi} \\ \widehat{\bm{\lambda}}_\psi
\end{bmatrix}
\right)
,
\end{align*} 
where $\widehat{\bm{\lambda}}_{\psi}$ 
is the maximum likelihood estimator of $\bm{\lambda}$
expressed in terms of the elements of
$\bm{\psi}$.

However, 
such approximation may be problematic. 
In fact, the profile log-likelihood is not 
a \emph{bona fide} log-likelihood function~\cite{BarndorffNielsen1994}. 
Moreover, 
the associated profile score 
and
the information biases are only guaranteed to be $O(1)$~\cite{SEVERINI01091998}. 
Therefore ML estimators based on profile log-likelihood
are prone to bias issues,
specially for small sample sizes~\cite{SilvaCribariFrery:ImprovedLikelihood:Environmetrics}.
%In other words, if $\widetilde{U}(\bm{\psi})=\partial \widetilde{\ell}(\bm{\psi})/\partial \bm{\psi}$, then
%$$
%\operatorname{E}[\widetilde{U}(\bm{\psi})]=O(1)
%$$ 
%and 
%$$
%\operatorname{E}[\widetilde{U}(\bm{\psi})\widetilde{U}(\bm{\psi})^t] + \operatorname{E}\bigg[\frac{\partial \widetilde{U}(\bm{\psi})}{\partial \bm{\psi}^t}\bigg]=O(1).
%$$

%Agora diga por que ser O(1) é ruim.
%However, there is a problem in the utilization of $\widetilde{\ell}(\cdot)$.
%The profile likelihood is not a genuine likelihood and the profile score and information biases are only guaranteed to be $O(1)$, i.e.,
%This fact implies that ML estimators based on profile log-likelihood can be more affected than usual log-likelihood for small value of $N$.
%In order to solve this problem, several authors have proposed modification for profile log-likelihood, such as Barndorff-Nielsen~\cite{BarndorffNielsen1994}, Cox and Reid~\cite{coxandreid1992}, McCullagh and Tibshirani~\cite{McCullaghandTibshirani1990}, and Stern~\cite{Stern1997}. 

To address this issue, 
several modifications for
the profile log-likelihood function
have been proposed~\cite{McCullaghandTibshirani1990,coxandreid1992,BarndorffNielsen1994,Stern1997}.
%This paper considered only the profile likelihood modified by Barndorff-Nielsen, which is described in the following discussion.
In the current work,
we adopted the Barndorff-Nielsen modification~\cite{BarndorffNielsen1994}, 
which is
analytically tractable 
when the scaled complex Wishart distribution is considered.

Barndorff-Nielsen approximation for the marginal log-likelihood function is expressed according to~\cite{BarndorffNielsen1983}:
\begin{align}
\label{l-tilde-bn}
\widetilde{\ell}_\text{BN}(\bm{\psi})
=
\widetilde{\ell}(\bm{\psi})
-
\log\
\left|
\frac{\partial \widehat{\bm{\lambda}}_\psi}{\partial\widehat{\bm{\lambda}} }
\right|
-
\frac{1}{2} 
\log
\left|
\mathcal{J}_{\bm{\lambda}\bm{\lambda}}
\left(
\begin{bmatrix}\bm{\psi} \\ \widehat{\bm{\lambda}}_\psi\end{bmatrix}
\right)
\right|
,
\end{align}
%\rjdsc{O que e' $\widehat{\bm{\lambda}}$?}
where 
$\widehat{\bm{\lambda}}$ 
is the unrestricted maximum likelihood estimators 
for the nuisance parameter 
$\bm{\lambda}$,
and
$\mathcal{J}_{\bm{\lambda}\bm{\lambda}} \left(\cdot\right)$ 
is the observed information matrix for $\bm{\lambda}$
given by:
\begin{align*}
\mathcal{J}_{\bm{\lambda}\bm{\lambda}}
\left(
\begin{bmatrix}\bm{\psi} 
\\ 
\widehat{\bm{\lambda}}_\psi\end{bmatrix}
\right)
=
-
\frac{\partial^2}{\partial \bm{\lambda} \partial \bm{\lambda}^\text{H}}
\ell(\bm{\theta})
.
\end{align*}
%
%When the $\bm{\lambda}$ of the above quantity is a complex vector, $\partial \bm{\lambda} \partial \bm{\lambda}^\top$ is replaced by $\partial \bm{\lambda} \partial \bm{\lambda}^\text{H}$.
%
The associated bias for $\widetilde{\ell}_\text{BN}(\bm{\psi})$
is $O(N^{-1})$~\cite{BarndorffNielsen1983}.
%
%Using the ``$p^*$ formula''
%In order to approximate the probability density function of ML estimator~\cite{BarndorffNielsen1983},  Barndorff-Nielsen proposed an approximation for marginal log-likelihood with bias of order $O(N^{-1})$ given by 
%$$
%\ell_\text{BN}(\bm{\psi})=\ell_p(\bm{\psi})-\log\biggl|\frac{\partial \widehat{\bm{\lambda}}_{\psi}}{\partial\widehat{\bm{\lambda}} }\biggr|-\frac 12 \log|\mathcal J_{\lambda\lambda}(\bm{\psi},\widehat{\bm{\lambda}}_{\psi})|,
%$$
%where $\mathcal J_{\lambda\lambda}(\bm{\psi},\widehat{\bm{\lambda}})=-\partial^2 \ell(\bm{\theta}) /\partial \bm{\lambda} \partial \bm{\lambda}^t$.
%When parameters are orthogonal 
%
%
If $\widehat{\bm{\lambda}}_\psi=\widehat{\bm{\lambda}}$ for all $\psi$
(i.e., 
${\partial \widehat{\bm{\lambda}}_{\psi}}/{\partial\widehat{\bm{\lambda}} }=\bm{I}_{p-q}$, 
where $\bm{I}_a$ represents a identity matrix with order $a$), 
\eqref{l-tilde-bn}
collapses into:
\begin{equation}
\label{bnlikelihod}
\widetilde{\ell}_\text{BN}(\bm{\psi})
=
\widetilde{\ell}(\bm{\psi})
-
\frac{1}{2}
\log
\left|
\mathcal{J}_{\lambda\lambda}
\left(
\begin{bmatrix}\bm{\psi} \\ \widehat{\bm{\lambda}}_\psi\end{bmatrix}
\right)
\right|
.
\end{equation}
Above expression is known as 
the Cox-Reid modified profile log-likelihood~\cite{coxandreid1992}.
%For situations in which the parameters are orthogonal (i.e., ${\partial \widehat{\bm{\lambda}}_{\psi}}/{\partial\widehat{\bm{\lambda}} }=1$), the above formula is degenerated in 
%\begin{equation}\label{bnlikelihod}
%\ell_\text{BN}(\bm{\psi})=\ell_p(\bm{\psi})-\frac 12 \log|\mathcal J_{\lambda\lambda}(\bm{\psi},\widehat{\bm{\lambda}}_{\psi})|,
%\end{equation}
%which is known as Cox and Reid modified profile log-likelihood~\cite{coxandreid1992}.
%\rjdsc{Confirmar a sentenca em sublinha.}
The vector $\widehat{\bm{\psi}}_\text{BN}$
that maximizes~\eqref{l-tilde-bn}
is the Barndorff-Nielsen ML estimator for~$\bm{\psi}$~\cite{SilvaCribariFrery:ImprovedLikelihood:Environmetrics}.

It is shown in~\cite{BarndorffNielsen1983}
that the bias of~$\widehat{\bm{\psi}}_\text{BN}$
is
$O(N^{-3/2})$.
Therefore, for
this particular log-likelihood function, 
the ML estimator bias is also
$O(N^{-3/2})$. 
This is an improvement over the standard ML estimator, 
whose bias is $O(N^{-1})$.
%~\cite{?}.

%Then we have that
%$$
%\operatorname{E}
%\left(
%\widehat{\bm{\psi}}_\text{BN}
%=
%\bm{\psi}
%+
%O(N^{-3/2}),
%$$
%i.e., $\widehat{\bm{\psi}}_{\text{BN}}$ is less biased than the usual ML estimator.

\section{Improved ML Estimation for the Scaled Complex Wishart 
Distribution}
\label{correction:results}

%Assuming the scaled complex Wishart distribution as an \textit{a priori} model,

%In~\cite{EstimationEquivalentNumberLooksSAR},
%Anfinsen~\textit{et al.}
%proposed 
%ML estimators
%for the parameters of the
%scaled complex Wishart distribution.
%
In this section
we propose
improvements to ML estimators 
for
the parameters of the
scaled complex Wishart distribution.
We
consider
the methodologies described in the previous
section,
namely 
the Cox-Snell corrected ML estimation
and
the Barndorff-Nielsen ML estimation based on the
modified profile log-likelihood function.

Mainly,
our goal is to derive
corrected ML estimators for
$\bm{\Sigma}$ and $L$.
However,
as established in~\cite{FreryCintraNascimento2012},
we notice that
$\operatorname{E}(\widehat{\bm{\Sigma}}_\text{ML}) = \bm{\Sigma}$.
Therefore,
$\widehat{\bm{\Sigma}}_\text{ML}$ is already
unbiased,
requiring no correction.
Thus,
we focus our efforts on obtaining
a corrected ML estimator for $L$.

\subsection{Cox-Snell Corrected ML estimators for $\bm{\Sigma}$ and $L$}

According to~\eqref{correctedestimator0},
the Cox-Snell method
yields
an improved ML estimator for $L$,
termed by $\widehat{L}_{\text{IML}}$,
given by
\begin{equation}
\widehat{L}_\text{IML}
=
\widehat{L}_\text{ML}
-
\widehat{B}(\widehat{L}_\text{ML})
.
\label{appestim}
\end{equation}

Now we aim at computing
$\widehat B(\widehat{L}_\text{ML})$.
For such firstly
we need to derive $B(\widehat{L}_{\text{ML}})$.
Invoking~\eqref{Bartlettcorrection}, we have that 
\begin{align}
B(\widehat{L}_\text{ML})=&\kappa^{L,L} \kappa^{L,L}\Bigl(\kappa_{LL}^{(L)}-\frac{1}{2} \kappa_{LLL}\Bigr)\nonumber \\
&+\sum_{i=1}^{m^2}\sum_{j=1}^{m^2} \kappa^{L,L} \kappa^{{\sigma}_i,{\sigma}_j}\Bigl(\kappa_{L{\sigma}_i}^{({\sigma}_i)}-\frac 12 \kappa_{L{\sigma}_i{\sigma}_j} \Bigr).
\label{nonumberbias}
\end{align}
Following algebraic manipulations, 
as detailed in Appendix~\ref{AP1}, 
we could simplify~\eqref{nonumberbias} and re-cast it as 
\begin{equation*}
B(L)=
\frac{m^2}{2 N L
\left[  
\psi_m^{(1)}( L ) -\frac{m}{L}
\right]}
-
\frac{\frac{m}{2 L}
+
\psi_m^{(2)}( L )}{2N
\left[  
\psi_m^{(1)}( L )  -\frac{m}{ L }
\right]^2
}
.
\end{equation*}

Now replacing $L$ by $\widehat{L}_\text{ML}$,
we obtain $\widehat B(\widehat{L}_\text{ML})$.
This quantity
must be inserted back into~\eqref{appestim}
to furnish
the proposed improved ML estimator for $L$.

\subsection{Modified profile log-likelihood for number of looks}
\label{profmodif}

Following 
the discussed Barndorff-Nielsen technique,
we 
derive the profile log-likelihood function
associated to the scaled complex Wishart distribution.
This quantity can be obtained by
replacing
the covariance matrix~$\bm{\Sigma}$
by its ML estimator~$\widehat{\bm{\Sigma}}_\text{ML}$
in~\eqref{likelihood}.
This particular manipulation yields
$\widetilde{\ell}(\cdot)$
as a function of the sought parameter~$L$:
\begin{equation} 
\label{wish:profile:log:like}
\begin{split}
\widetilde{\ell}(L)
=
&
mNL(\log L-1)+(L-m)\sum_{k=1}^N\log|\bm{Z}_k| 
\\
&
-NL\log|\bar{\bm{Z}}|-N\frac{m(m-1)}{2}\log\pi 
\\
&
-N\sum_{k=0}^{m-1}\log\Gamma(L-k)
.
\end{split}
\end{equation}
Thus, the profile score function
is given by
\begin{align*}
\widetilde{U}(L)
=
\frac{\partial \widetilde{\ell}(L)}{\partial L}
=
&
m N \log L - N\psi_m^{(0)}(L) - N\log|\bar{\bm{Z}}|
\\
&
+ \sum_{k=1}^N\log|\bm{Z}_k|
.
\end{align*}

%\rjdsc{(6) requer que os parametros seja 'ortogonais'. Demonstrar que sao ortogonais em appendix.}
%\ADCN{ Na verdade, o fato do estimador irrestrito ser similar ao estimador restrito resulta na obten\c{c}\~ao de (6).
% Adicionalmente, este fato tamb\'em implica na ortigonalidade de par\^ametro.
% Fiz uma modifica\c{c}\~ao da frase acima da express\~ao (6) neste sentido.}

From~\eqref{likelihood}, 
we could express the profile score function in terms of 
$\bm{\sigma}=\operatorname{vec}(\bm{\Sigma})$,
yielding:
\begin{align*}
U(\bm{\sigma})
=
N
\operatorname{vec}(
\bm{\Sigma}^{-1}
\bar{\bm{Z}}
\bm{\Sigma}^{-1}
-
\bm{\Sigma}^{-1}
)
.
\end{align*}
By imposing $U(\bm{\sigma})=\bm{0}$,
where $\mathbf{0}$ is a column vector of zeros,
and solving the resulting system of equations,
one obtains 
that 
the
unrestricted and restricted ML estimators
for $\bm{\sigma}$
are given by
$\operatorname{vec}(\bar{\bm{Z}})$.
Moreover,
we have that
%\rjdsc{$\widehat{\bm{\sigma}}_L$ parece nunca ter sido definido.}
%\ADCN{Feito}
%
\begin{align*}
\frac{\partial \widehat{\bm{\sigma}}_L}
{\partial \widehat{\bm{\sigma}}}
=
\bm{I}_{m^2},
\end{align*}
where $\widehat{\bm{\sigma}}_L$ represents the restricted ML estimator.
As a consequence,
%\eqref{l-tilde-bn} collapses into~\eqref{bnlikelihod}.
we are now in a position to use~\eqref{bnlikelihod}.
%\rjdsc{Nao entendo esta sentenca; melhorar.}
% adopting $L$ as interest parameter and $\bm{\sigma}$ as the nuisance parameter
%vector in the scaled complex Wishart distribution.

However,
the following quantity is necessary:
\begin{align*}
-\frac 12 \log\left|
\mathcal{J}_{\bm{\sigma}\bm{\sigma}}
\left(
\begin{bmatrix}
L \\ \widehat{\bm{\sigma}}
\end{bmatrix}
\right)
\right|
,
\end{align*}
where
$\widehat{\bm{\sigma}}
= 
\operatorname{vec}(\widehat{\bm{\Sigma}}_\text{ML})$.
Due to its length,
the mathematical derivation for this quantity
is detailed in~Appendix~\ref{AP2}.
Thus,
from~\eqref{bnlikelihod},
the modified profile log-likelihood 
according to the Barndorff-Nielsen technique
is furnished by
\begin{align}
\widetilde{\ell}_{\text{BN}}(L)
=
\widetilde{\ell}(L)
-
\frac{m^2}{2}
(\log N+\log L)
-
m\log|{\bar{\bm{Z}}}^{-1}|
.
\label{log:like:BN}
\end{align}
As a consequence,
the associated
relative score function is expressed by 
\begin{equation*}
\widetilde{U}_\text{BN}(L)
=
\frac{\partial}{\partial L}
\widetilde{\ell}_\text{BN}(L)
=
\widetilde{U}(L)
-
\frac{m^2}{2L}
.
%\label{BNestimator}
\end{equation*}
%respectively.

As required,
the sought estimator~$\widehat{L}_\text{BN}$
must satisfy
$\widetilde{U}_{\text{BN}}(\widehat{L}_\text{BN})=0$.
In other words,
it is the solution of the following
non-linear equation:
\begin{equation}
\label{eq-L-hat-BN}
\begin{split}
%\widetilde{U}(\widehat{L}_\text{BN})-\frac{m^2}{2\widehat{L}_\text{BN}}=0.
m N
\log \widehat{L}_\text{BN} 
&
-
N\psi_m^{(0)}
(\widehat{L}_\text{BN})
-
N\log|\bar{\bm{Z}}|
\\
&
+\sum_{k=1}^N
\log|\bm{Z}_k|
-
\frac{m^2}{2\widehat{L}_\text{BN}}
=
0
.
\end{split}
\end{equation}
A closed-form expression for $\widehat{L}_\text{BN}$ 
could not be given.
Then we resort to numerical techniques,
such as the Newton-Raphson method,
as a means for solving~\eqref{eq-L-hat-BN}.

\section{Performance and Assessment}
\label{correction:resultsII}

To assess the performance of the proposed estimators
for the equivalent number of looks,
we considered a simulation study on synthetic data
based on Monte Carlo experiments.
Such simulation
included the
following
estimators:
\begin{enumerate}
\item 
the standard ML estimator 
($\widehat{L}_\text{ML}$)~\cite{EstimationEquivalentNumberLooksSAR};

\item 
the first and second trace moment estimators 
($\widehat{L}_\text{MM1}$ and $\widehat{L}_\text{MM2}$, respectively)~\cite{EstimationEquivalentNumberLooksSAR};

\item 
the proposed ML estimator based on the Cox and Snell methodology
($\widehat{L}_\text{IML}$);
and

\item 
the proposed ML estimator according to the Barndorff-Nielsen adjustment
($\widehat{L}_\text{BN}$).

\end{enumerate}
The first three above-mentioned estimators
were derived by Anfinsen~\textit{et al.}~\cite{EstimationEquivalentNumberLooksSAR}.
%\ADCN{Comentei a seguinte sentença, pois os estimadores baseados em momentos tinham sido definidos na seção~II-B. }
%The estimators based 
%on trace moments
%are given by 
%%
%\begin{align*}
%\widehat{L}_\text{MM1}
%=
%\frac{\operatorname{tr}(\overline{\bm{Z}}\;\overline{\bm{Z}})}
%{N^{-1}\sum_{i=1}^N \operatorname{tr}(\bm{Z}_i)^2-\operatorname{tr}(\overline{\bm{Z}})^2}
%\end{align*}
%and
%\begin{align*}
%\widehat{L}_\text{MM2}
%=
%\frac{\operatorname{tr}(\overline{\bm{Z}})^2}
%{N^{-1}\sum_{i=1}^N \operatorname{tr}(\bm{Z}_i\bm{Z}_i)-\operatorname{tr}(\overline{\bm{Z}}\;\overline{\bm{Z}})}.
%\end{align*}

Simulation results,
were assessed
in terms of 
three statistical measures:
(i)~the mean squared error (MSE);
(ii)~the coefficient of variation (CV);
and
(iii)~the mean estimated values;
and
(iv)~the bias curves ($B$).

Subsequently
we separate the best estimator among those proposed 
by Anfinsen~\textit{et al.}
as indicated 
by the simulation study.
Then
such selected estimator 
as well as 
the proposed estimators
are submitted
to actual data analysis
aiming at
performance assessment.

%As the three latter estimators were already studied in~\cite{EstimationEquivalentNumberLooksSAR}, we analyze them only on synthetic scenarios which were not considered.
%

\subsection{Simulation study}

%This section aims to compare the proposed estimators to others three derived and explored by Anfinsen~\textit{et al.}~\cite{EstimationEquivalentNumberLooksSAR}.
%That paper proposed two estimators based on trace moment whose expressions are given by 
%$$
%\widehat{L}_\text{MM1}=\frac{\operatorname{tr}(\overline{\bm{Z}}\;\overline{\bm{Z}})}{N^{-1}\sum_{i=1}^N \operatorname{tr}(\bm{Z}_i)^2-\operatorname{tr}(\overline{\bm{Z}})^2}
%$$
%and
%$$
%\widehat{L}_\text{MM2}=\frac{\operatorname{tr}(\overline{\bm{Z}})^2}{N^{-1}\sum_{i=1}^N \operatorname{tr}(\bm{Z}_i\bm{Z}_i)-\operatorname{tr}(\overline{\bm{Z}}\;\overline{\bm{Z}})}.
%$$
%%where $\overline{\operatorname{tr}(\bm{Z})^2}=N^{-1}\sum_{i=1}^N \operatorname{tr}(\bm{Z}_i)^2$ and $\overline{\operatorname{tr}(\bm{Z}\bm{Z})}=N^{-1}\sum_{i=1}^N \operatorname{tr}(\bm{Z}_i\bm{Z}_i)$.

%Moreover, a rich approach was performed concerning the ML estimator obtained numerically by the solution of~\eqref{eqscore1}. 
%In the following discussion, these estimators are compared to two new proposals in both synthetic and actual study.
%In summary, the considered estimators are 
%\begin{enumerate}
%\item standard ML estimator ($\widehat{L}_\text{ML}$)~\cite{EstimationEquivalentNumberLooksSAR},
%\item improved ML estimator based on the Cox and Snell methodology ($\widehat{L}_\text{IML}$),
%\item improved ML estimator according to the Barndorff-Nielsen adjustment ($\widehat{L}_\text{BN}$), and
%\item first and second trace moment estimators ($\widehat{L}_\text{MM1}$ and $\widehat{L}_\text{MM2}$, respectively).
%\end{enumerate}

The suggested Monte Carlo simulation employed
5500~replications as discussed 
in~\cite{HypothesisTestingSpeckledDataStochasticDistances}.
Additionally,
the following simulation parameters
were considered:
(i)~sample sizes $N\in \{9,49,121\}$;
(ii)~number of looks $L\in\{4,6,8,12\}$;
and
(iii)~a covariance matrix given by
$$ 
\bm{\Sigma}_0
=
\left[ 
\begin{array}{ccc}
962892&19171-3579\textbf{i}&-154638+191388\textbf{i} \\
&56707&-5798+ 16812\textbf{i} \\
&&472251
\end{array} 
\right]
,
$$
where the omitted elements can be obtained 
from the conjugation of their respective symmetric elements.

The adopted sample sizes are associated
to square windows of $\{3,7,11\}$
pixels, respectively.
It is worth mentioning that
higher values of $L$ represent images
less affected by speckle noise.
This particular estimated covariance matrix~$\bm{\Sigma}_0$
was obtained by the E-SAR sensor over We{\ss}ling, Germany.
In particular,
it was employed in~\cite{PolarimetricSegmentationBSplinesMSSP}
for the characterization of urban areas.

%presents the average of estimates for $L$ as well as the respective values for the comparison criteria.
%

The results for MSE, 
CV, 
and
mean estimated values are shown in Table~\ref{tableapplication3}.
Initially
we notice that
the proposed estimators
$\widehat{L}_\text{BN}$
and
$\widehat{L}_\text{IML}$
could offer
more accurate
average ML estimates
for 
the ENL
when
compared to the results 
derived from the other considered methods.
In general terms, 
the increasing of sample size reduces the MSE and CV.
This behavior is expected, 
because the estimators are asymptotically unbiased.
Results also show that the estimator 
$\widehat{L}_\text{IML}$ 
presented the best performance 
in terms of MSE and CV
in all considered cases.
This last result provides even stronger evidence 
in favor of the accuracy of $\widehat{L}_\text{IML}$. 
This is because 
bias corrected estimators do not always yield
better procedures than ML estimators.
Indeed, 
bias corrected estimators
can offer larger variance values,
leading to increased MSE values
when compared to uncorrected methods~\cite{Cordeiro2011}.

%This last result provides even stronger evidence in favor of the accuracy of $\widehat{L}_\text{IML}$, 
%because bias corrected estimators do not always yield better procedures than ML estimators, 
%since they sometimes have larger variance leading to larger MSE than the uncorrected procedures~\cite{Cordeiro2011}.

\begin{figure*}[htbp] 
\centering
\subfigure[$L=4$\label{Bias1}]{\includegraphics[width=.48\linewidth]{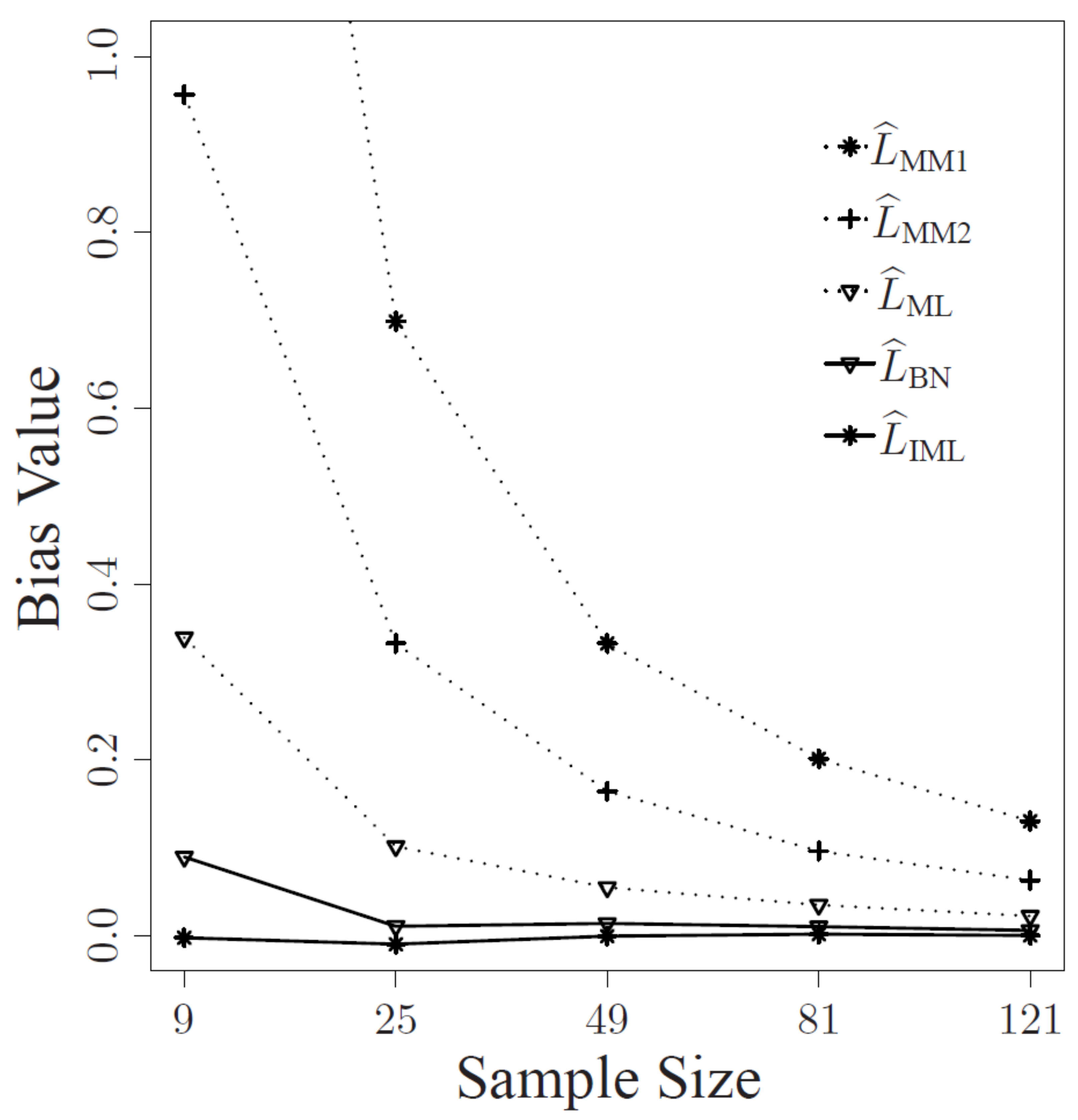}}
\subfigure[$L=6$\label{Bias22}]{\includegraphics[width=.48\linewidth]{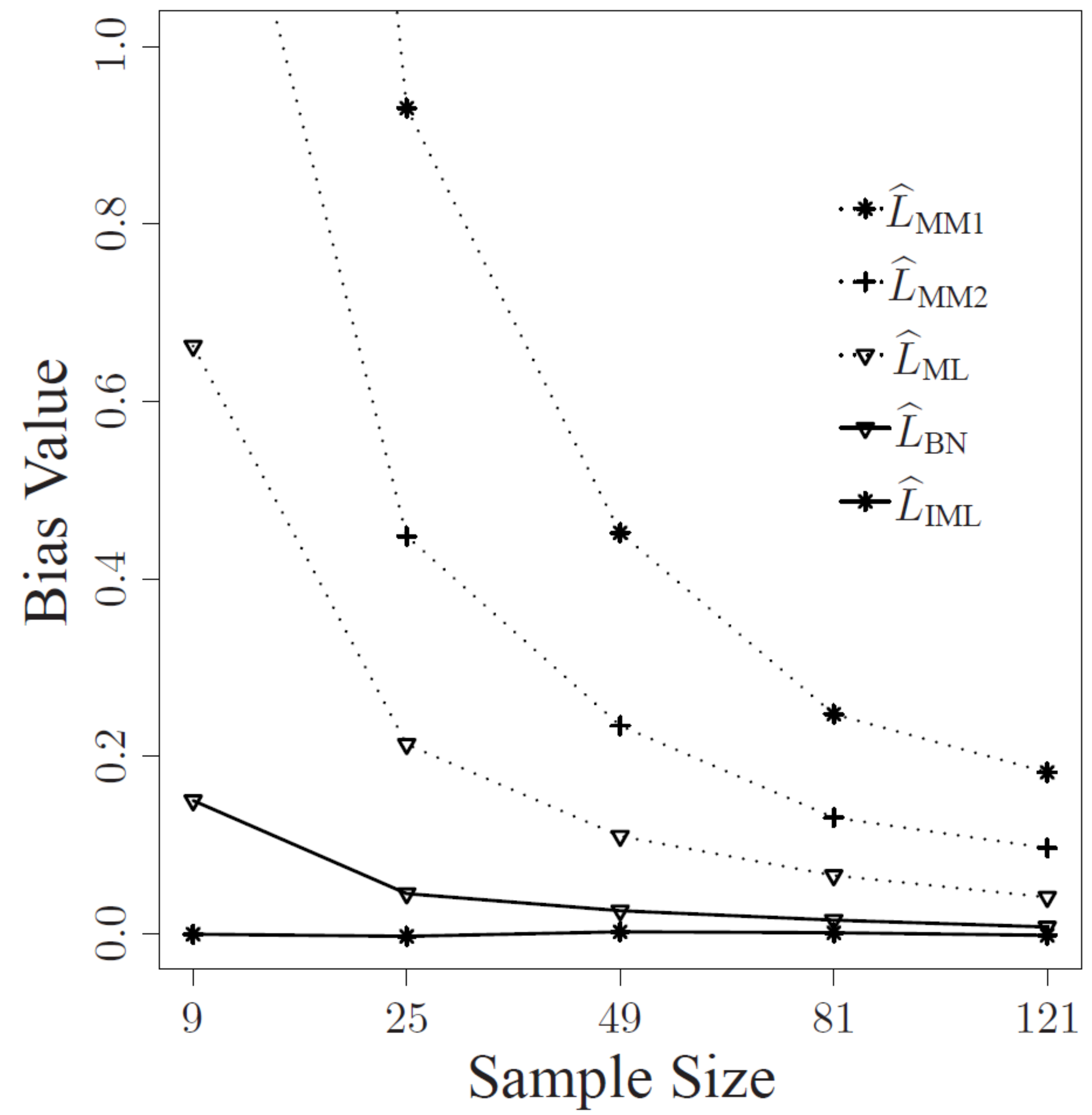}}\\
\subfigure[$L=8$\label{Bias3}]{\includegraphics[width=.48\linewidth]{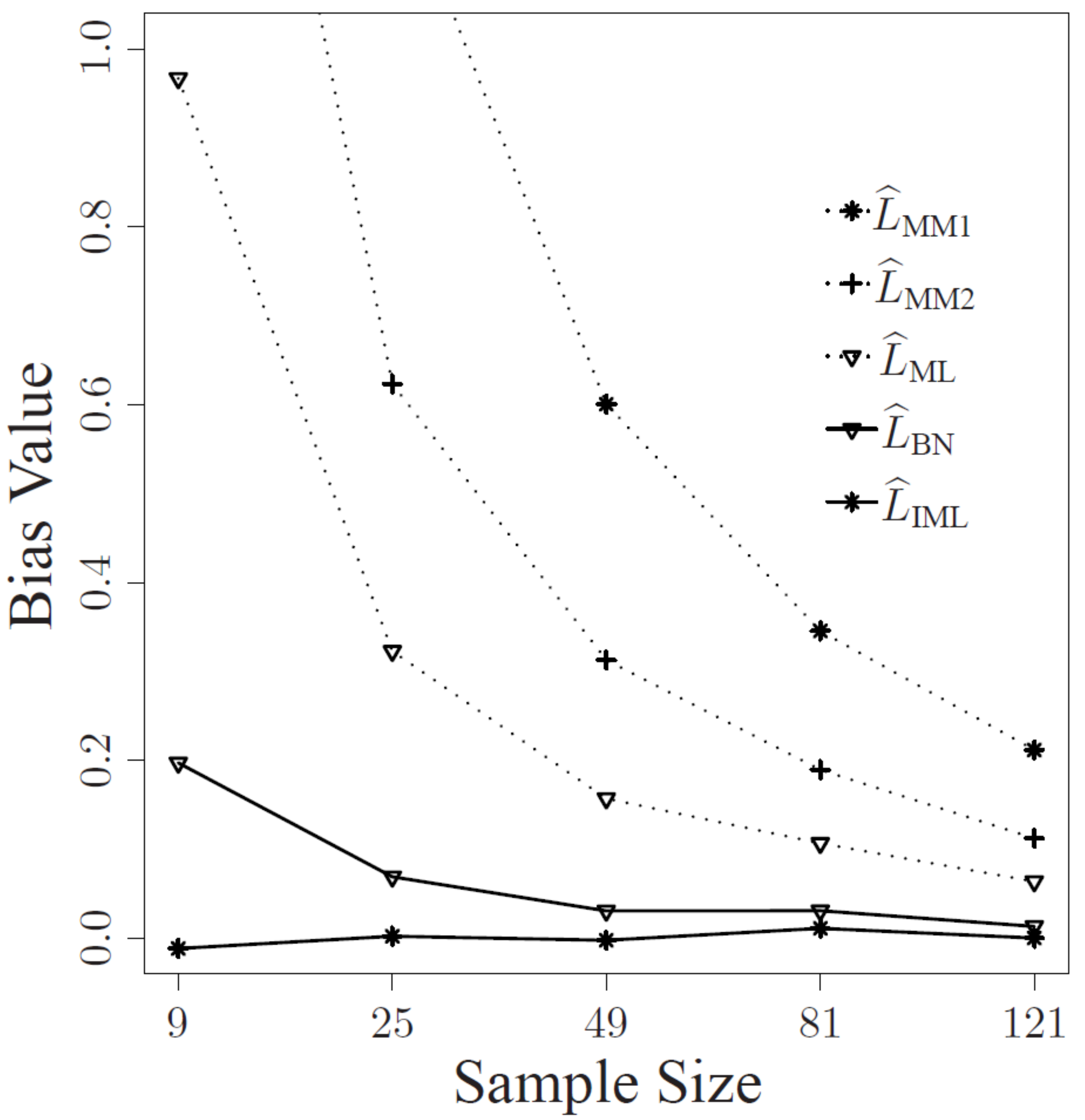}}
\subfigure[$L=12$\label{Bias4}]{\includegraphics[width=.48\linewidth]{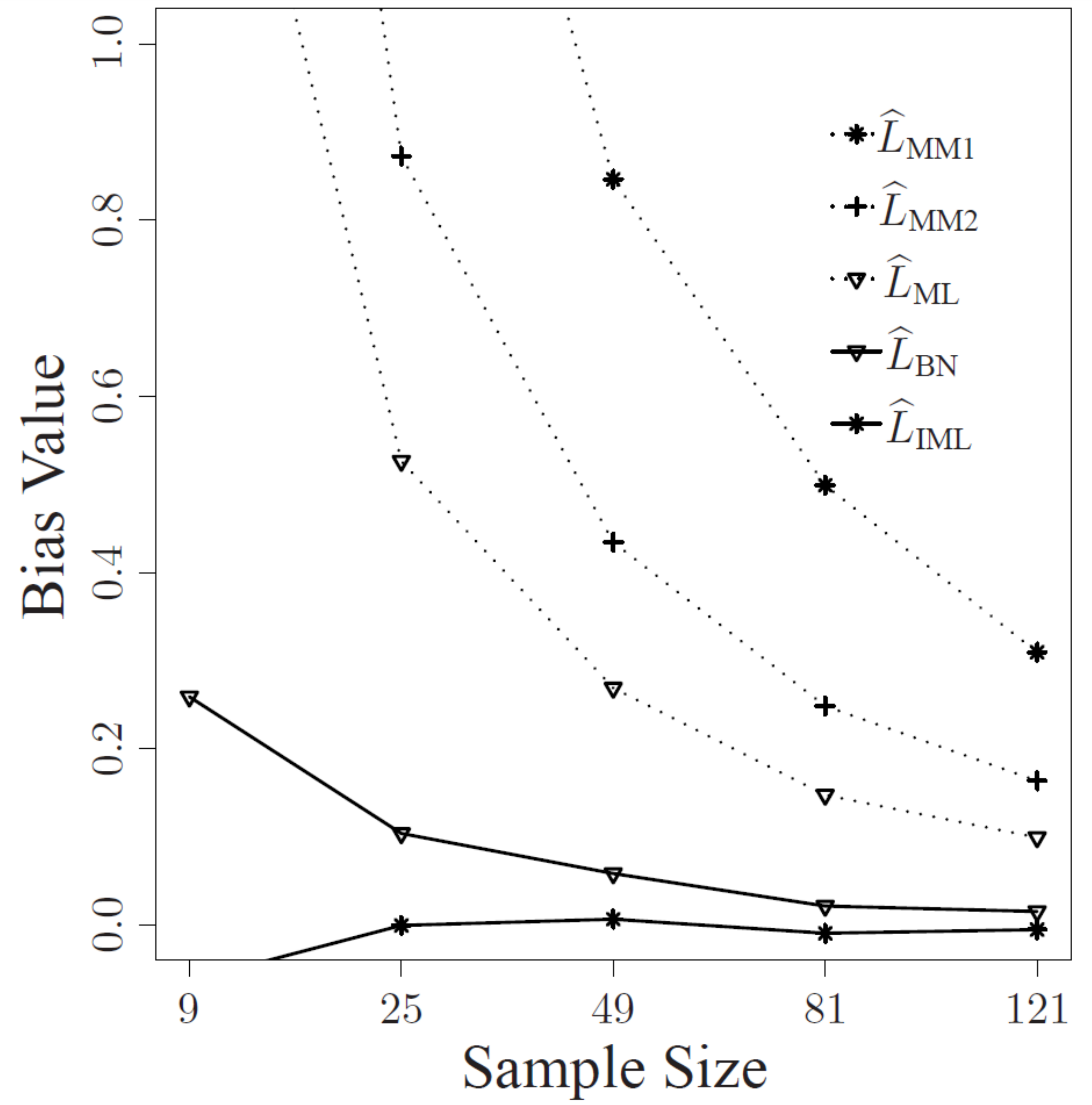}}
\caption{Bias curves in terms of sample sizes for synthetic data.}
\label{Bias}
\end{figure*}

Fig.~\ref{Bias} presents the values of bias for several sample sizes.
The following inequality is verified:
\begin{align*}
\widehat{B}(\widehat{L}_\text{MM1})
\geq
\widehat{B}(\widehat{L}_\text{MM2})
\geq
\widehat{B}(\widehat{L}_\text{ML})
\geq
\widehat{B}(\widehat{L}_\text{BN})
\geq
\widehat{B}(\widehat{L}_\text{IML})
.
\end{align*}
It is noteworthy that, with the exception of $\widehat{L}_\text{IML}$, all the procedures overestimate the equivalent number of looks, i.e., they lead to decisions that assume that there is more information in the data than the true content.
The only estimator with a different behavior is $\widehat{L}_\text{IML}$.
It consistently exhibits the smallest bias, which is reduced with increasing number of looks.
For large $L$ and small samples, its bias becomes negative.
The other estimators have their biases increased with the number of looks.

The estimator $\widehat{L}_\text{IML}$ was also less affected
by the increasing of number of looks.
Thus, 
this estimator performed well
enduring both variations of the sample size
as well as the number of looks.   
We also emphasize
that
the better performance of 
estimator
$\widehat{L}_\text{ML}$
in comparison
with
$\widehat{L}_\text{MM1}$ 
and
$\widehat{L}_\text{MM2}$
is in agreement with the findings of Anfisen~\emph{et al.}~\cite{EstimationEquivalentNumberLooksSAR}.
However,
when considering the proposed estimators,
we identify that
$\widehat{L}_\text{ML}$,
$\widehat{L}_\text{IML}$,
and
$\widehat{L}_\text{BN}$
as the more accurate estimators
among the considered techniques.

\begin{table*}[hbtp]
\centering
\footnotesize
\caption{Measures of the estimators performance with synthetic data}
\label{tableapplication3}
%\begin{tabular}{cc rrr rrr rrr rrr}\toprule                                                                           
\begin{tabular}{c@{ }c@{ }  r@{ }r@{ }r@{\quad}  r@{ }r@{ }r@{\quad} r@{ }r@{ }r@{\quad} r@{ }r@{ }r}\toprule
\multicolumn{2}{c}{}& \multicolumn{3}{c}{$L=4$} & \multicolumn{3}{c}{$L=6$} & \multicolumn{3}{c}{$L=8$}&
\multicolumn{3}{c}{$L=12$} \\                            
\cmidrule(lr{.25em}){1-2}\cmidrule(lr{.25em}){3-5} \cmidrule(lr{.25em}){6-8} 
\cmidrule(lr{.25em}){9-11} \cmidrule(lr{.25em}){12-14}
$\widehat{L}$ & $N$
& \multicolumn{1}{c}{$\operatorname{mean}(\widehat{L})$} & 
\multicolumn{1}{c}{$\operatorname{MSE}$} & \multicolumn{1}{c}{$\operatorname{CV}$} 
& \multicolumn{1}{c}{$\operatorname{mean}(\widehat{L})$} & 
\multicolumn{1}{c}{$\operatorname{MSE}$} & \multicolumn{1}{c}{$\operatorname{CV}$} 
& \multicolumn{1}{c}{$\operatorname{mean}(\widehat{L})$}  & 
\multicolumn{1}{c}{$\operatorname{MSE}$} & \multicolumn{1}{c}{$\operatorname{CV}$}
& \multicolumn{1}{c}{$\operatorname{mean}(\widehat{L})$} & 
\multicolumn{1}{c}{$\operatorname{MSE}$} & \multicolumn{1}{c}{$\operatorname{CV}$}  \\
\cmidrule(lr{.25em}){1-2}\cmidrule(lr{.25em}){3-5} \cmidrule(lr{.25em}){6-8} 
\cmidrule(lr{.25em}){9-11} \cmidrule(lr{.25em}){12-14}
$\widehat{L}_\text{ML}$ & 9                                                                                    
& 4.339	&	0.414	 &	0.126	&	6.663	&	1.373	 &	0.145 &	8.967	 &	2.810	 &	0.153	&	13.538	&	6.963   &	0.158 \\ 
$\widehat{L}_\text{MM1}$ &                                                                                     
& 6.278	&	23.174 &	0.676	&	9.344	&	52.605 &	0.689 &	12.478 &	95.978 &	0.698	&	18.119	&	190.432 &	0.683 \\ 
$\widehat{L}_\text{MM2}$ &                                                                                     
& 4.957	&	2.993	 &	0.291	&	7.401	&	6.399	 &	0.285 &	9.849	 &	11.800 &	0.294	&	14.606	&	24.977  &	0.292 \\ 
$\widehat{L}_\text{IML}$ &                                                                                     
& 3.998	&	0.221	 &	0.118	&	6.000	&	0.695	 &	0.139 &	7.989	 &	1.398	 &	0.148	&	11.937	&	3.435   &	0.155 \\ 
$\widehat{L}_\text{BN}$ &                                                                                      
& 4.090	&	0.243	 &	0.118	&	6.150	&	0.760	 &	0.140 &	8.197	 &	1.518	 &	0.148	&	12.259	&	3.700   &	0.155 \\ 
                                                                                                               
\cmidrule(lr{.25em}){1-2}\cmidrule(lr{.25em}){3-5} \cmidrule(lr{.25em}){6-8} \cmidrule(lr{.25em}){9-11}        
\cmidrule(lr{.25em}){12-14}                                                                                    
                                                                                                               
&49& 4.055 &	0.042	&	0.049	&	6.110	&	0.133 &	0.057	&	8.157	&	0.258 &	0.059	&	12.269	&	0.661 &	0.063	\\       
  && 4.333 &	1.045	&	0.223	&	6.452	&	2.201 &	0.219	&	8.601	&	3.809 &	0.216	&	12.847	&	8.363 &	0.215	\\       
  && 4.165 &	0.294	&	0.124	&	6.235	&	0.633 &	0.122	&	8.313	&	1.093 &	0.120	&	12.435	&	2.388 &	0.119	\\       
  && 4.000 &	0.037	&	0.048	&	6.002	&	0.115 &	0.056	&	7.998	&	0.222 &	0.059	&	12.007	&	0.559 &	0.062	\\       
  && 4.014 &	0.037	&	0.048	&	6.026	&	0.117 &	0.057	&	8.031	&	0.225 &	0.059	&	12.059	&	0.568 &	0.062	\\       
                                                                                                               
\cmidrule(lr{.25em}){3-5} \cmidrule(lr{.25em}){6-8} \cmidrule(lr{.25em}){9-11} \cmidrule(lr{.25em}){12-14}     
                                                                                                               
&121& 4.023	&	0.016	&	0.031 &	6.041	&	0.048 &	0.036	&	8.064	&	0.096 &	0.038	&	12.100	&	0.237 &	0.039 \\       
   && 4.131	&	0.350	&	0.140 &	6.182	&	0.738 &	0.136	&	8.212	&	1.261 &	0.134	&	12.310	&	2.820 &	0.134 \\       
   && 4.063	&	0.108	&	0.080 &	6.097	&	0.233 &	0.077	&	8.113	&	0.403 &	0.077	&	12.164	&	0.871 &	0.076 \\       
   && 4.001	&	0.015	&	0.031 &	5.998	&	0.045 &	0.035	&	8.001	&	0.090 &	0.038	&	11.995	&	0.222 &	0.039 \\       
   && 4.006	&	0.015	&	0.031 &	6.008	&	0.045 &	0.035	&	8.014	&	0.091 &	0.038	&	12.016	&	0.223 &	0.039 \\       
\bottomrule
\end{tabular}
\end{table*}

\subsection{Analysis with actual data}

\begin{figure}[htb]
\centering
\includegraphics[width=.8\linewidth]{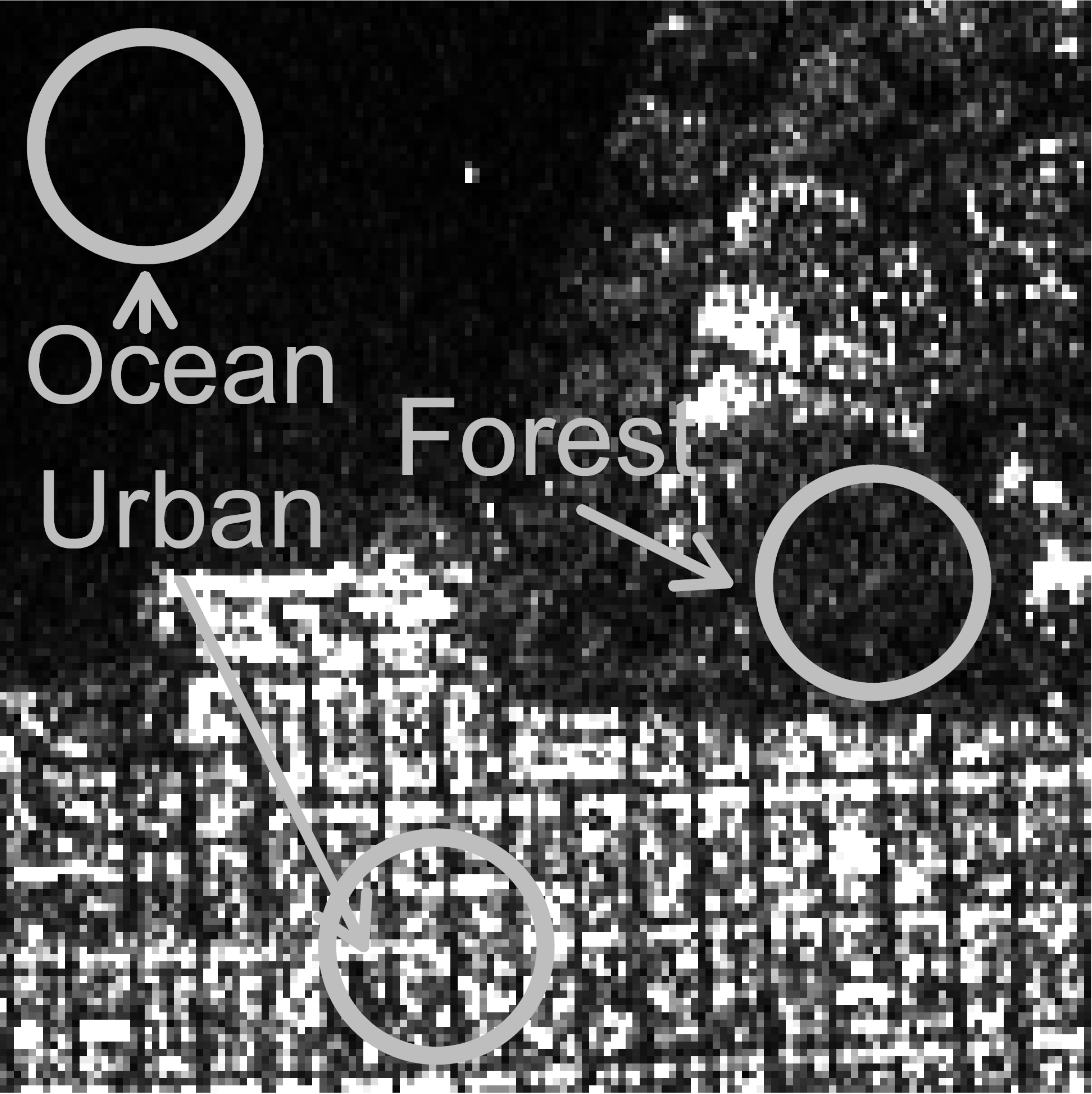}
\caption{San Francisco image with selected regions.}
\label{SanFransicoim}
\end{figure}

Now we aim at submitting actual PolSAR data
to the three best
estimators separated in the previous subsection.
%the three better assessed estimators as identifies under the synthetic data analysis.
%
%
We adopted
the San Francisco AIRSAR image
as a source of actual data.
This classical multilook PolSAR image 
was obtained by the AIRSAR sensor operating at L~band
with four nominal looks~\cite{ESA2009}.
Fig.~\ref{SanFransicoim} presents 
the HH channel intensity data of the $150\times150$ San Francisco image.
Sub-areas were selected from the main image.
By means of visual inspection,
three categories were sought: 
(i)~ocean;
(ii)~forest; 
and
(iii)~urban area.

For each selected subregion, 
we considered 5500~subsamples 
without replacement with size
$N\in\{9,36,121,144\}$.
These subsamples are then submitted to parameter estimation.

Table~\ref{tableapplication4} displays the obtained results.
For the selected ocean subregion, 
the corrected estimator $\widehat{L}_\text{IML}$ 
presented the smallest values of MSE and CV.
Thus,
the furnished estimates were more accurate.
As discussed 
by Anfinsen~\textit{et al.}~\cite{EstimationEquivalentNumberLooksSAR}, 
textured regions yield to underestimation effects.
Then correction schemes tend to become less efficient 
when compared to usual ML estimation $\widehat{L}_\text{ML}$. 
%\ADCN{
%Additionally, Freitas~\textit{et al.}~\cite{FreitasFreryCorreia:Environmetrics:03}
%showed that the intensification of texture in PolSAR images
%becomes the utilization of the complex Wishart
%distribution inappropriate.  
%%
%Notice that this phenomenon has been reported
%in the literature~\cite{FreryCorreiaFreitas:ClassifMultifrequency:IEEE:2007,BombrunBeaulieu2008,BombrunVasileGayTotir2011}.
%%
%However, other polarimetric
%distributions to describing textured areas present intractable
%expressions which depend on special functions, such as the
%hypergeometric and modified Bessel functions. 
%%
%And, therefore,
%modest progresses have been made on their statistical models
%in the context of ML estimation and much less on improving this kind of estimator in small sample sizes.}
%

Indeed, 
as reported in~\cite{FreitasFreryCorreia:Environmetrics:03}, 
heavily textured images, 
such as
forest and urban, 
are less adequately described by the complex Wishart distribution.
This phenomenon was also verified in~\cite{FreryCorreiaFreitas:ClassifMultifrequency:IEEE:2007,BombrunBeaulieu2008,BombrunVasileGayTotir2011}.
%In fact,
%the current literature still lacks a sufficiently flexible model to encompass widely 
%varying texture levels.
Therefore, 
the proposed tools are expected to excel 
at low textured samples.

%In Table~\ref{tableapplication4},
%we summarize our results
%which are in agreement with literature findings. 
%%
%The following discussion is
%focused on the better modeled ocean data.

\begin{table*}[hbt]
\centering 
\footnotesize
\caption{Measures of the estimators performance with actual data}\label{tableapplication4}   
\begin{tabular}{c@{ }c@{ }  r@{\,\,\,}r@{\,\,\,}r@{\,\,\,}  r@{\,\,\,}r@{\,\,\,}r@{\,\,\,} r@{\,\,\,}r@{\,\,\,}r@{\,\,\,}}\toprule
\multicolumn{2}{c}{}& \multicolumn{3}{c}{$\widehat{L}_\text{ML}$} & \multicolumn{3}{c}{$\widehat{L}_\text{IML}$} & \multicolumn{3}{c}{$\widehat{L}_\text{BN}$} \\
\cmidrule(lr{.25em}){1-2}\cmidrule(lr{.25em}){3-5} \cmidrule(lr{.25em}){6-8}
\cmidrule(lr{.25em}){9-11}
Regions & $N$ & \multicolumn{1}{c}{$\operatorname{mean}(\widehat{L})$} &
\multicolumn{1}{c}{$\operatorname{MSE}$} & \multicolumn{1}{c}{$\operatorname{CV}$} 
& \multicolumn{1}{c}{$\operatorname{mean}(\widehat{L})$} &
\multicolumn{1}{c}{$\operatorname{MSE}$} & \multicolumn{1}{c}{$\operatorname{CV}$} 
& \multicolumn{1}{c}{$\operatorname{mean}(\widehat{L})$}  &
\multicolumn{1}{c}{$\operatorname{MSE}$} & \multicolumn{1}{c}{$\operatorname{CV}$}\\
\cmidrule(lr{.25em}){1-2}\cmidrule(lr{.25em}){3-5} \cmidrule(lr{.25em}){6-8}
\cmidrule(lr{.25em}){9-11}

Ocean                                                                                 
& 9   & 4.520	&	0.747 &	0.153  & 3.776	&	0.350 &	0.145  &	3.995	&	1.376 &	0.294	\\  
& 36  & 4.074	&	0.623 &	0.193  & 3.932	&	0.582 &	0.193  &	3.979	&	0.739 &	0.216	\\  
& 121 & 4.140	&	0.034 &	0.029  & 4.099	&	0.024 &	0.029  &	4.123	&	0.029 &	0.029	\\  
& 144 & 4.133	&	0.029 &	0.026  & 4.099	&	0.021 &	0.025  &	4.118	&	0.025 &	0.026	\\  
\cmidrule(lr{.25em}){1-2}\cmidrule(lr{.25em}){3-5} \cmidrule(lr{.25em}){6-8}          
\cmidrule(lr{.25em}){9-11}                                                            
Forest                                                                                
&& 3.293	&	0.599 &	0.095   &	3.069	 &	0.938 &	0.087  &	3.159		&	0.792  &	0.092 \\
&& 3.119	&	0.794 &	0.043   &	3.075	 &	0.872 &	0.042  &	3.089		&	0.852  &	0.049 \\
&& 2.809	&	2.041 &	0.281   &	2.798	 &	2.065 &	0.281  &	2.818		&	1.987  &	0.273 \\
&& 3.082	&	0.845 &	0.017   &	3.072	 &	0.864 &	0.017  &	3.075		&	0.858  &	0.017 \\
\cmidrule(lr{.25em}){1-2}\cmidrule(lr{.25em}){3-5} \cmidrule(lr{.25em}){6-8}          
\cmidrule(lr{.25em}){9-11}                                                            
Urban                                                                                 
&& 2.738	&	2.439 &	0.336 &	2.492	 &	3.239 &	0.394  &	2.678	&	2.451 &	0.313	    \\  
&& 2.567	&	2.486 &	0.256 &	2.529	 &	2.594 &	0.260  &	2.454	&	2.989 &	0.316	    \\  
&& 2.218	&	3.959 &	0.399 &	2.209	 &	3.990 &	0.401  &	2.177	&	4.146 &	0.417	    \\  
&& 2.447	&	2.833 &	0.265 &	2.439	 &	2.857 &	0.265  &	2.426	&	2.923 &	0.275	    \\  

\bottomrule
\end{tabular}
\end{table*}

\begin{figure}[htb]
\centering
\includegraphics[width=\linewidth]{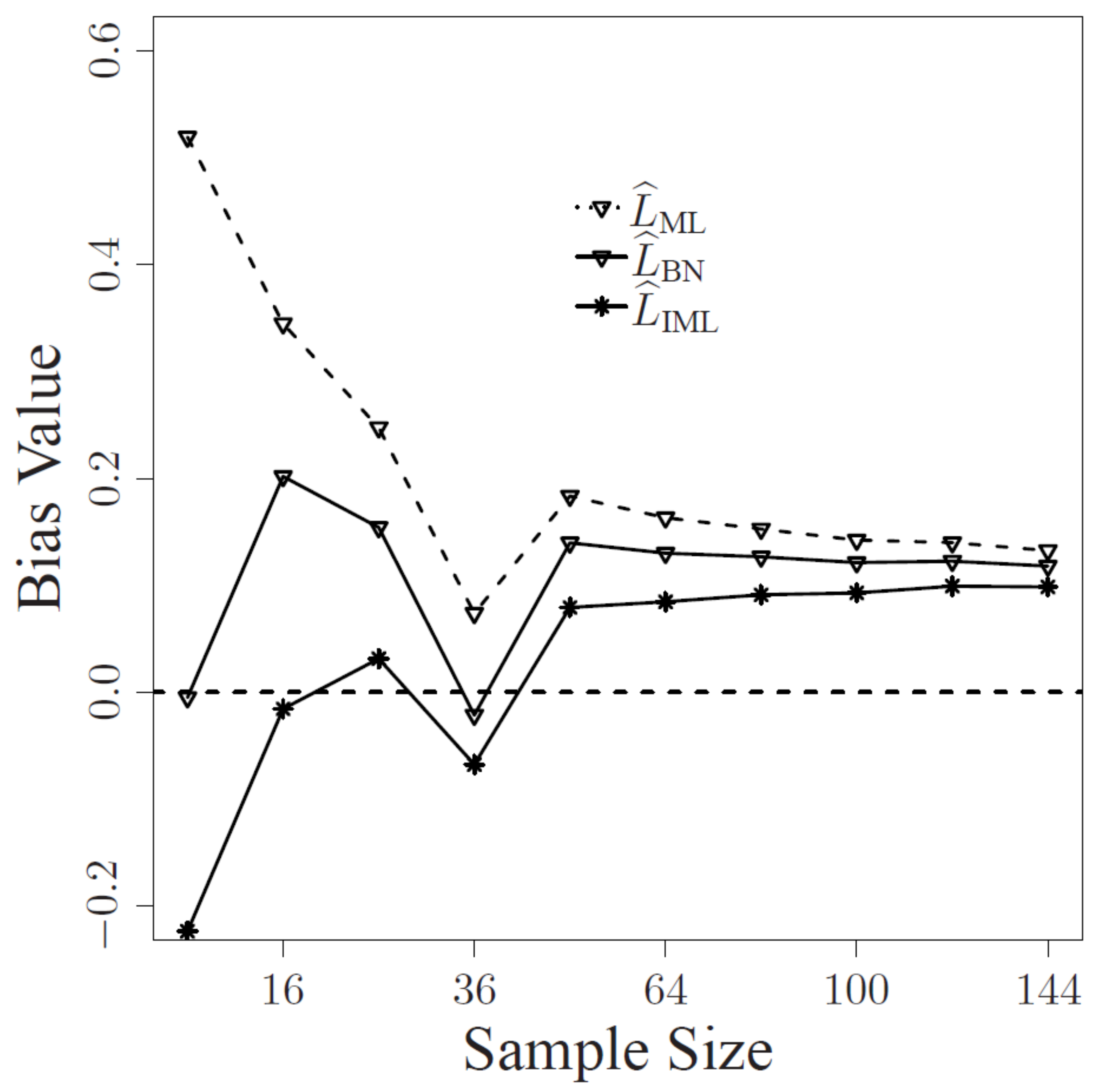}
\caption{Bias curves in terms of sample sizes for actual data.}
\label{Bias2}
\end{figure}

Fig.~\ref{Bias2} shows
the bias curves 
for the ocean region
in which case
the proposed tools were superior.
These curves indicate that
the estimator $\widehat{L}_\text{IML}$ 
could outperform
the usual ML estimator
when homogeneous targets are considered.

%As expected, results showed evidence that the estimators tend to produce  
%the same asymptotic behavior as from a given sample size.
  
Finally,
we notice
that both Table~\ref{tableapplication4} and Figure~\ref{Bias2}
indicate that
the discussed estimators
possess
similar asymptotic properties.
This behavior is expected and
can derived from theoretical results
as shown in~\cite{caselaberge2002,Kay1998}.

\section{Conclusion}
\label{correction:conclusion}

%This paper has suggested a advance in the ML estimation method for the parameters of the scaled complex 

In this paper, 
we proposed two new improved corrected ML estimators
for the scaled complex Wishart distribution.
Emphasis was put on the estimation of the number of looks.
In particular,
we introduced closed-form expressions 
for the  bias and for the modified profile likelihood
according to
methodologies advanced by 
Cox and Snell~\cite{CoxandSnell1968} 
and 
Barndorff-Nielsen~\cite{BarndorffNielsen1994}.  

By means of Monte Carlo simulations,
the proposed techniques
were assessed and compared with
estimators presented by 
Anfinsen~\textit{et al.}~\cite{EstimationEquivalentNumberLooksSAR}.
Adopted figures of merit include
MSE, 
coefficient of variation, 
mean of estimated values,
and the bias curves.
Results showed that the proposed corrected estimators
outperform all considered estimators.
Actual data were also analyzed
revealing that the proposed estimators 
are superior
when homogeneous scenes are considered.
 
\appendices

\section{Cumulants for the Cox-Snell correction}\label{AP1}

In this appendix, we present closed-form expressions for (i)~some cumulants from the scaled complex Wishart distribution as well as (ii)~the second-order bias according to the Cox-Snell approach.

From~\eqref{likelihood}, the second and third-order derivatives are given by
\begin{align}
U_{LL}=&\,\,\frac{Nm}{L}-N\psi_m^{(1)}(L),\label{IF:1} \\
U_{L{\sigma}_i}=&\,\,N[\operatorname{vec}(\bm{\Sigma}^{-1}\bar{\bm{Z}}\bm{\Sigma}^{-1}-\bm{\Sigma}^{-1})]_i,\label{IF:1-1} \\
U_{{\sigma}_i{\sigma}_j}=&\,\,LN[(\bm{\Sigma}^{-1} \otimes \bm{\Sigma}^{-1})-(\bm{\Sigma}^{-1} \otimes \bm{\Sigma}^{-1})\bar{\bm{Z}}\bm{\Sigma}^{-1}\nonumber \\
&\quad\quad-(\bm{\Sigma}^{-1} \otimes \bm{\Sigma}^{-1})\bm{\Sigma}^{-1}\bar{\bm{Z}}]_{i,\;j},\label{IF:2} \\ 
U_{L{\sigma}_i{\sigma}_j}=&\,\,L^{-1}U_{{\sigma}_i{\sigma}_j},\nonumber \\
U_{{\sigma}_i{\sigma}_j{\sigma}_k}=&\,\,LN[3(\bm{\Sigma}^{-1} \otimes \bm{\Sigma}^{-1})\otimes(\bm{\Sigma}^{-1}\bar{\bm{Z}}\bm{\Sigma}^{-1})\nonumber \\
&+3(\bm{\Sigma}^{-1} \otimes \bm{\Sigma}^{-1})\otimes(\bm{\Sigma}^{-1}\bm{\Sigma}^{-1}\bar{\bm{Z}})\nonumber \\
&-2(\bm{\Sigma}^{-1} \otimes \bm{\Sigma}^{-1})\otimes \bm{\Sigma}^{-1}
 ]_{k+(i-1)m^2,\;k+(j-1)m^2},\nonumber
\end{align}
for $i,k,j=1,2,\ldots,m^2$, where $\otimes$ is the Kronecker product and ${\sigma}_i$ is the $i$th element of vector $\bm{\sigma}$. 

Based on the above results, we derive the following expression for the cumulants:

\begin{align}
\kappa^{L,L}=&\,\,\frac 1N\bigl[\psi_m^{(1)}(L)-\frac{m}{L}\bigr]^{-1}, \nonumber \\
\kappa^{L,{\sigma}_i}=&\,\,0,\quad \kappa^{{\sigma}_i,{\sigma}_j}=\frac{1}{NL}(\bm{\Sigma} \otimes \bm{\Sigma})_{i,\;j}, \nonumber \\
\kappa_{LLL}=&\,\,\kappa_{LL}^{(L)}=-N\Bigl[\frac{m}{L^2}+ \psi_m^{(2)}(L)\Bigr], \nonumber \\
\kappa_{LL{\sigma}_i}=&\,\,\kappa_{LL}^{({\sigma}_i)}=\kappa_{L{\sigma}_i}^{({\sigma}_j)}=0, \label{eqref1} \\
\kappa_{{\sigma}_i{\sigma}_jL}=&\,\,-N(\bm{\Sigma}^{-1} \otimes \bm{\Sigma}^{-1})_{i,\;j}, \nonumber \\
\kappa_{{\sigma}_i{\sigma}_j}^{({\sigma}_k)}=&\,\,2LN \nonumber \\
&[(\bm{\Sigma}^{-1} \otimes \bm{\Sigma}^{-1})\otimes \bm{\Sigma}^{-1}%\nonumber \\
]_{k+(i-1)m^2,\;k+(j-1)m^2},\nonumber \\%\label{eqref2}\\
\kappa_{{\sigma}_i{\sigma}_j{\sigma}_k}=&\,\,4LN [(\bm{\Sigma}^{-1} \otimes \bm{\Sigma}^{-1})\otimes \bm{\Sigma}^{-1}
]_{k+(i-1)m^2,\;k+(j-1)m^2}. \nonumber 
%\label{eqref3}
\end{align}

Since $\kappa^{L{\sigma}_i}=0$, parameters $L$ and ${\sigma}_i$, for $i=1,2,\ldots,m^2$, are orthogonal~\cite{CoxandReid1987}.
Therefore, expression~\eqref{Bartlettcorrection} yields:
\begin{align*}
B(\widehat{L}_\text{ML})=&\kappa^{L,L} \kappa^{L,L}\Bigl(\kappa_{LL}^{(L)}-\frac{1}{2} \kappa_{LLL}\Bigr)\\
&+\sum_{i=1}^{m^2}\sum_{j=1}^{m^2} \kappa^{L,L} \kappa^{{\sigma}_i,{\sigma}_j}\Bigl(\kappa_{L{\sigma}_i}^{({\sigma}_i)}-\frac 12 \kappa_{L{\sigma}_i{\sigma}_j} \Bigr).
\end{align*}
Above formula can be given a closed form expression. 
Indeed, from~\eqref{eqref1}, we have that
\begin{align*}
\kappa^{{\sigma}_i,{\sigma}_j}\Bigl(\kappa_{L{\sigma}_i}^{({\sigma}_i)}&-\frac 12 \kappa_{L{\sigma}_i{\sigma}_j} \Bigr)\\
&=\frac{1}{NL}(\bm{\Sigma} \otimes \bm{\Sigma})_{i,\;j}\bigl(0+ \frac N2\{\bm{\Sigma}^{-1} \otimes \bm{\Sigma}^{-1}\}_{i,\;j}\bigr)\\
&=\frac{1}{2L} (\bm{\Sigma} \otimes \bm{\Sigma})_{i,\;j}\{(\bm{\Sigma} \otimes \bm{\Sigma})^{-1}\}_{i,\;j}.
\end{align*}
Considering the general expression of the elements of the inverse matrix~\cite[pp.~82]{johnson1988applied}, we have:
$$
\sum_{i=1}^{m^2}\sum_{j=1}^{m^2}\{\bm{\Sigma} \otimes \bm{\Sigma}\}_{i,\;j}\{(\bm{\Sigma} \otimes \bm{\Sigma})^{-1}\}_{i,\;j}=m^2.
$$
Therefore, we obtain:
\begin{equation*}
B(\widehat{L}_{\text{ML}})=
\frac{m^2}{2NL\bigl[  \psi_m^{(1)}(L) -\frac{m}{L}\bigr]}-  \frac{\frac{m}{2L}+  \psi_m^{(2)}(L)   }{2N\bigl[  \psi_m^{(1)}(L)  -\frac{m}{L}\bigr]^2}.
%\label{expresscoxsnell}
\end{equation*}
%Therefore, the first new estimator is given by
%$$
%\widehat{L}_{\text{IML}}=\widehat{L}_{\text{ML}}-\widehat{B}(\widehat{L}_{\text{ML}}).
%$$

\section{Derivation for Barndorff-Nielsen approximation}\label{AP2}

In the subsequent discussion, we aim to detail the derivation of the Barndorff-Nielsen approximation for the marginal log-likelihood function of the scaled complex Wishart distribution.
In particular, this appendix addresses the derivation for the quantity 
$
-\frac 12 \log\left|
\mathcal{J}_{\bm{\sigma}\bm{\sigma}}
\left(
\begin{bmatrix}L \\ \widehat{\bm{\sigma}}\end{bmatrix}
\right)
\right|
$
 discussed in Subsection~\ref{profmodif}.

Based on results \eqref{IF:1}-\eqref{IF:2}, one can rewrite the observed information matrix from the scaled complex Wishart distribution by the following matrix equation:
\begin{align*}
&\hskip+12ex\Bigg[ \begin{array}{cc}
U_{LL}& U^\top_{L\operatorname{vec}(\boldsymbol{\Sigma})} \\
U_{\operatorname{vec}(\boldsymbol{\Sigma})^* L} & U_{\operatorname{vec}(\boldsymbol{\Sigma})\operatorname{vec}(\boldsymbol{\Sigma})^*}
\end{array} \Bigg]=\\
%&\\
&\Bigg[ \begin{array}{cc}
\frac{mN}{L} - N\psi_m^{(1)}(L)& N\operatorname{vec}(\boldsymbol{\Sigma}^{-1}\bar{\boldsymbol{Z}}\boldsymbol{\Sigma}^{-1}-\boldsymbol{\Sigma}^{-1})^{\top} \\
N\operatorname{vec}(\boldsymbol{\Sigma}^{-1}\bar{\boldsymbol{Z}}\boldsymbol{\Sigma}^{-1}-\boldsymbol{\Sigma}^{-1})^* & U_{\operatorname{vec}(\boldsymbol{\Sigma}) \operatorname{vec}(\boldsymbol{\Sigma})^*} 
\end{array} \Bigg],
\end{align*}
where $U^\top_{L\operatorname{vec}(\boldsymbol{\Sigma})}$, defined by
$$
U^\top_{L\operatorname{vec}(\boldsymbol{\Sigma})}= \frac{\partial^2}{\partial L \partial \operatorname{vec}(\boldsymbol{\Sigma})^\top} \ell(\bm{\theta}),
$$
is a vector whose the $i$th-element is given in~\eqref{IF:1-1} and~\cite{EstimationEquivalentNumberLooksSAR} 
%$U_{\operatorname{vec}(\boldsymbol{\Sigma})\operatorname{vec}(\boldsymbol{\Sigma})^*}$, given by
\begin{align*}
U_{\operatorname{vec}(\boldsymbol{\Sigma})\operatorname{vec}(\boldsymbol{\Sigma})^*}&= \frac{\partial^2}{\partial \operatorname{vec}(\boldsymbol{\Sigma})^* \partial \operatorname{vec}(\boldsymbol{\Sigma})^\top }\ell(\bm{\theta})\\
&=LN[(\bm{\Sigma}^{-1} \otimes \bm{\Sigma}^{-1})-(\bm{\Sigma}^{-1} \otimes \bm{\Sigma}^{-1})\bar{\bm{Z}}\bm{\Sigma}^{-1}\\
&\hskip+7ex-(\bm{\Sigma}^{-1} \otimes \bm{\Sigma}^{-1})\bm{\Sigma}^{-1}\bar{\bm{Z}}].
\end{align*}
%represents a matrix with ($i,j$)-entry defined in~\eqref{IF:2}.

Thus, the quantity 
$
\mathcal{J}_{\bm{\sigma}\bm{\sigma}}
\left(
\begin{bmatrix}L \\ \widehat{\bm{\sigma}}\end{bmatrix}
\right)
$
%$\mathcal{J}_{\bm{\lambda}\bm{\lambda}}$ in~\eqref{l-tilde-bn} can be replaced by
can be defined by
\begin{align*}
%\mathcal{J}_{\operatorname{vec}(\boldsymbol{\Sigma})\operatorname{vec}(\boldsymbol{\Sigma})^*}
%\left(
%\begin{bmatrix}L \\ \operatorname{vec}(\widehat{\boldsymbol{\Sigma}}_{\text{ML}})^\top
%\end{bmatrix}
%\right)
\mathcal{J}_{\bm{\sigma}\bm{\sigma}}
\left(
\begin{bmatrix}L \\ \widehat{\bm{\sigma}}\end{bmatrix}
\right)
&=
%\mathcal{J}_{\operatorname{vec}(\boldsymbol{\Sigma})\operatorname{vec}(\boldsymbol{\Sigma})^*}
%\left(
%\begin{bmatrix}L \\ \operatorname{vec}(\overline{\boldsymbol{Z}})^\top
%\end{bmatrix}
%\right)\\
-U_{\operatorname{vec}(\boldsymbol{\Sigma})\operatorname{vec}(\boldsymbol{\Sigma})^*}\Big|_{\bm{\Sigma}=\bar{\bm{Z}}}\\
&=
NL \bar{\boldsymbol{Z}}^{-1} \otimes \bar{\boldsymbol{Z}}^{-1}.
\end{align*}
and, therefore, the following quantity combined with profile log-likelihood~\eqref{wish:profile:log:like} at~\eqref{bnlikelihod} yields the Barndorff-Nielsen approximation for the marginal log-likelihood function:
\begin{align}
&-\frac 12 \log\left|
%
%\mathcal{J}_{\operatorname{vec}(\boldsymbol{\Sigma})\operatorname{vec}(\boldsymbol{\Sigma})^*}
%\left(
%\begin{bmatrix}L \\ \operatorname{vec}(\widehat{\boldsymbol{\Sigma}}_{\text{ML}})^\top
%\end{bmatrix}
%\right)
%
\mathcal{J}_{\bm{\sigma}\bm{\sigma}}
\left(
\begin{bmatrix}L \\ \widehat{\bm{\sigma}}\end{bmatrix}
\right)
\right|\nonumber \\
&\hskip+7ex=-\frac{m^2}{2}(\log N+\log L)-\frac 12 {\log\bigl|\bar{\bm{Z}}^{-1} \otimes \bar{\bm{Z}}^{-1}\bigr|}.
\label{lastformula}
\end{align}
From~\cite[result 11.1~(l), page~235]{Seber:2007:MHS:1564898},
we have that
$$
\log\bigl|\bar{\bm{Z}}^{-1} \otimes \bar{\bm{Z}}^{-1}\bigr|={2m\log|{\bar{\bm{Z}}}^{-1}|}.
$$
Thus, \eqref{lastformula} can be simplified according to:
\begin{align*}
&-\frac 12 \log\left|
%
%\mathcal{J}_{\operatorname{vec}(\boldsymbol{\Sigma})\operatorname{vec}(\boldsymbol{\Sigma})^*}
%\left(
%\begin{bmatrix}L \\ \operatorname{vec}(\widehat{\boldsymbol{\Sigma}}_{\text{ML}})^\top
%\end{bmatrix}
%\right)
%
\mathcal{J}_{\bm{\sigma}\bm{\sigma}}
\left(
\begin{bmatrix}L \\ \widehat{\bm{\sigma}}\end{bmatrix}
\right)
\right|\nonumber \\
&\hskip+9ex=-\frac{m^2}{2}(\log N+\log L)-m\log|{\bar{\bm{Z}}}^{-1}|
.
\end{align*}
Thus,
we have that \eqref{log:like:BN} holds true.

\bibliographystyle{IEEEtran}
\bibliography{bibtexart} 

%[ADCN:] Atualizei o ``status'' do meu perfil.
\begin{IEEEbiography}[{\includegraphics[width=1in]{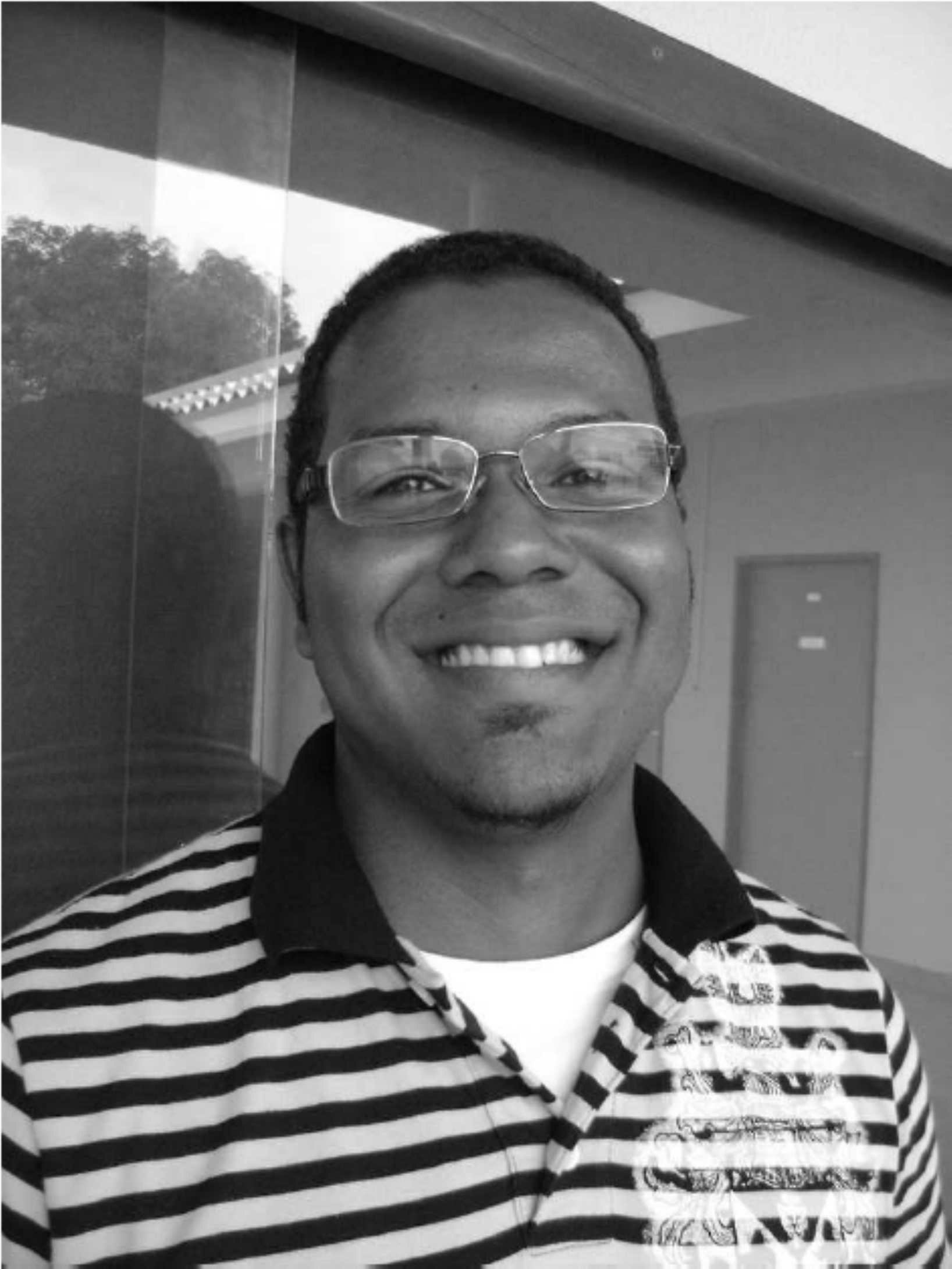}}]{Abra\~ao D.\ C.\ Nascimento}
holds B.Sc.\, M.Sc.\, and D.Sc. degrees in Statistics from Universidade Federal de Pernambuco (UFPE), Brazil, in 2005, 2007, and 2012, respectively.
In 2013, he joined the Department of Statistics at UFPB as Adjoint Professor.
His research interests are statistical information theory, inference on random matrices, and asymptotic theory.
\end{IEEEbiography}

\begin{IEEEbiography}[{\includegraphics[width=1in]{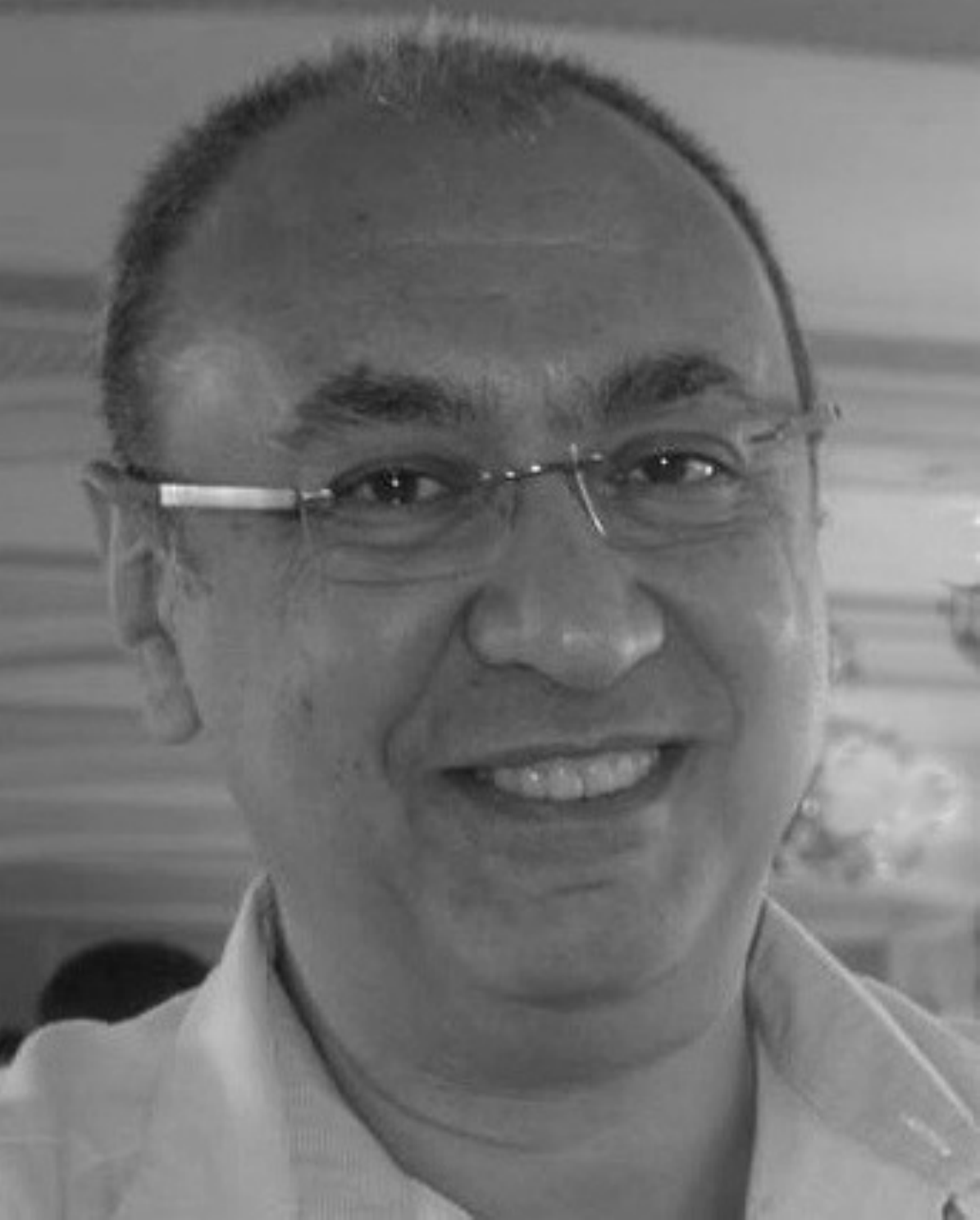}}]{Alejandro C.\ Frery} (S'92--SM'03)
received the B.Sc. degree in Electronic and Electrical Engineering from the Universidad de Mendoza, Mendoza, Argentina.
His M.Sc. degree was in Applied Mathematics (Statistics) from the Instituto de Matem\'atica Pura e Aplicada (IMPA, Rio de Janeiro) and his Ph.D. degree was in Applied Computing from the Instituto Nacional de Pesquisas Espaciais (INPE, S\~ao Jos\'e dos Campos, Brazil).
He is currently the leader of LaCCAN -- \textit{Laborat\'orio de Computa\c c\~ao Cient\'ifica e An\'alise Num\'erica}, Universidade Federal de Alagoas, Macei\'o, Brazil.
His research interests are statistical computing and stochastic modelling.
\end{IEEEbiography}

\begin{IEEEbiography}[{\includegraphics[width=1in]{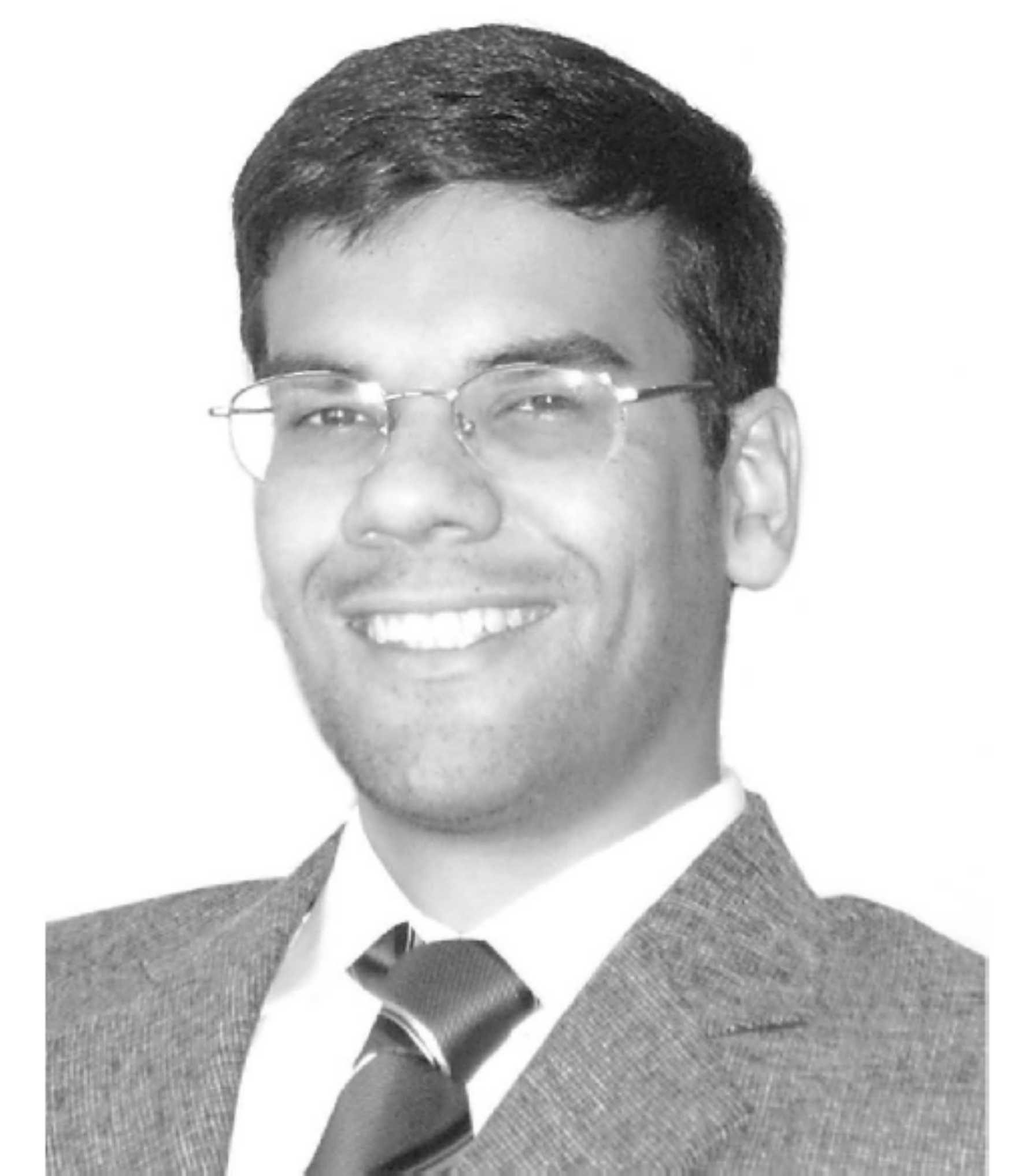}}]{Renato J.\ Cintra} (SM'10)
earned his B.Sc., M.Sc., and D.Sc. degrees
in Electrical Engineering from
Universidade Federal de Pernambuco,
Brazil, in 1999, 2001, and 2005, respectively.
In 2005,
he joined the Department of Statistics at UFPE.
During 2008-2009,
he worked at the University of Calgary, Canada,
as a visiting research fellow.
He is also a graduate faculty member of the
Department of Electrical and Computer Engineering,
University of Akron, OH.
His long term topics of research include
theory and methods for digital signal processing,
communications systems, and applied mathematics.
\end{IEEEbiography}

\vfill

\end{document}